\documentclass{article} 
\usepackage{iclr2027_conference,times}

\usepackage{amsmath,amsfonts,bm}

\def\eqref#1{equation~\ref{#1}}

\def\1{\bm{1}}

\DeclareMathAlphabet{\mathsfit}{\encodingdefault}{\sfdefault}{m}{sl}
\SetMathAlphabet{\mathsfit}{bold}{\encodingdefault}{\sfdefault}{bx}{n}

\usepackage{hyperref}
\usepackage{url}
\usepackage{xcolor}
\usepackage{graphicx}
\usepackage{booktabs}
\usepackage{pifont}
\usepackage{multirow}
\usepackage[table]{xcolor}
\usepackage{amsthm}

\title{MOVE: Multimodal Open-world Verification and Expansion for Graph Learning}

\author{
\normalfont
Zekai Chen$^{1}$ \quad
Jiayang Xing$^{1}$ \quad
Xun Wu$^{1}$ \quad
Zhang Miao$^{1}$ \quad
Xunkai Li$^{1}$ \\
Kairui Yang$^{1}$ \quad
Zhengyu Wu$^{1}$ \quad
Xu Wang$^{2}$ \quad
Rong-Hua Li$^{1}$ \quad
Guoren Wang$^{1}$ \\[1ex]
$^{1}$Beijing Institute of Technology \\
$^{2}$Shandong University \\
\texttt{Email of Zekai Chen:zackchen02@163.com}
}

\iclrfinalcopy
\begin{document}

\maketitle

\begin{abstract}
Multimodal graph learning faces a fundamental challenge: new classes may emerge after deployment, while models are trained with a fixed label space. Existing approaches typically detect unknown nodes and use LLMs to generate candidate class descriptions, but they do not determine whether existing classes are insufficient to cover these nodes or whether a generated class is reliable enough to expand the class space. Our empirical study reveals three challenges: multimodal information beyond individual modalities is required for unknown-node identification, LLM-generated class descriptions may not fully capture multimodal class characteristics, and directly adding candidate classes can introduce redundant categories. Based on these observations, we propose MOVE, a multimodal open-world class verification and expansion framework. MOVE identifies nodes that cannot be assigned to existing classes by jointly considering visual tokens, textual attributes, and graph context, leverages a multimodal LLM to generate candidate classes, and selectively expands the class space only when candidates are consistently supported by multimodal evidence without introducing unnecessary categories. Experiments demonstrate that MOVE achieves an average improvement of 11.87\% across unknown recognition, open-domain annotation, and downstream graph learning tasks.
\end{abstract}

\section{Introduction}
\label{sec:introduction}

Multimodal graphs provide a natural representation for real-world systems such as e-commerce platforms, social networks, and multimedia recommendation. However, these systems continuously evolve after deployment: new products, topics, and user interests may emerge beyond the classes observed during training. This motivates open-world multimodal graph learning (MM-OGL), where models need to identify nodes that cannot be assigned to existing classes and determine whether the class space should be expanded. Recent advances in open-world graph learning (OGL), such as OGA~\citep{wen2025llms}, move beyond unknown-node detection by assigning semantic labels to discovered nodes and incorporating newly identified classes into the model. Yet, for multimodal graphs, a natural-language class description alone is insufficient: the class of a node may also depend on visual tokens and its surrounding graph context, which may provide information unavailable from text alone. Therefore, the key problem is not simply generating a class name, but determining whether the visual, textual, and structural information of a node can still place it into an existing class, and whether a newly proposed class is sufficiently supported by these multimodal signals to justify expanding the class space.

To understand where this process fails, we conduct an empirical study following the OGL pipeline in Sec.~\ref{sec:empirical_study}. \textbf{\ding{172} Multimodal Assignment Failure.} The first challenge is determining whether a node truly belongs to an unseen class. A node may share similar visual or textual characteristics with known classes, while considering its visual tokens, textual tokens, and graph context together reveals that it cannot be assigned to any existing class (Sec.~\ref{sec:empirical_gap}). This indicates that unknown discovery should consider more than similarity in a single modality. \textbf{\ding{173} Ambiguous Class Induction.} Once such nodes are identified, the next challenge is determining what class they represent. Different visual patterns, textual descriptions, and neighboring nodes may provide different clues, so a multimodal LLM can produce multiple plausible class descriptions (Sec.~\ref{sec:empirical_proposal}). Such a description is only a candidate: before being treated as a new class, it should be mapped back to the multimodal representation space and checked against the visual, textual, and structural information of the corresponding nodes. \textbf{\ding{174} Unreliable Class Expansion.} However, even a candidate supported by the current nodes should not be directly added to the model. It may only fit the observations used to generate it or unnecessarily split an existing class, causing redundant categories when the class space is expanded (Sec.~\ref{sec:empirical_expansion}).

These observations suggest that MM-OGL should be viewed as a continuous process from identifying nodes outside the current class space, to inferring what class they may belong to, and finally deciding whether that class should be added. The key is that each decision should use the same multimodal information rather than relying on a language label alone: visual and textual tokens together with graph context determine whether a node can be assigned to an existing class, while a proposed class must also be supported by these signals before it is incorporated.

Motivated by this perspective, we propose \textbf{MOVE} (\textbf{M}ultimodal \textbf{O}pen-world Class \textbf{V}erification and \textbf{E}xpansion), which couples the above three decisions in a unified framework. MOVE first performs \textbf{Uncertainty-Aware Multimodal Coding} to identify nodes that cannot be assigned to existing classes by jointly modeling visual tokens, textual tokens, and graph context. It then conducts \textbf{Evidence-Grounded Class Induction}, where a multimodal LLM proposes candidate class descriptions and each description is transformed into the shared multimodal space to determine whether it is supported by the corresponding nodes. Finally, \textbf{Selective Class Expansion} evaluates whether the supported candidate should be introduced into the class space, retaining only candidates that remain effective beyond the observations used to generate them while avoiding redundant classes and interference with existing ones.

\textbf{Our Contributions.}
(1) \textbf{\underline{\textit{New Perspective.}}}
We reveal a progressive failure chain in MM-OGL, from identifying nodes beyond existing classes, to ambiguous class annotation and unreliable class expansion. We formulate the problem as determining when the current class space becomes insufficient and when a new class should be introduced.
(2) \textbf{\underline{\textit{New Framework.}}}
We propose MOVE, a MM-OGL framework that unifies unknown-node identification, candidate class induction, and selective class expansion, with multimodal LLMs serving only as candidate generators rather than final decision makers.
(3) \textbf{\underline{\textit{SOTA Performance.}}}
Experiments on four multimodal attributed graph benchmarks and two downstream tasks show that MOVE achieves the best performance and improves unknown-concept semantic similarity by 11.87\% on average.

\section{Problem Formulation}
\label{sec:problem}

We consider a multimodal attributed graph (MAG)
$\mathcal{G}=(\mathcal{V},\mathcal{E},\mathbf{X}^{v},\mathbf{X}^{t})$,
where each node is characterized by visual and textual attributes together
with graph topology. During training, only a subset
$\mathcal{V}_{l}\subset\mathcal{V}$ is labeled, whose labels belong to the
known concept set $\mathcal{C}^{K}$. After deployment, the remaining nodes
$\mathcal{V}_{u}=\mathcal{V}\setminus\mathcal{V}_{l}$ may contain both
known and previously unseen concepts:
\begin{equation}
    \mathcal{V}_{u}
    =
    \mathcal{V}_{u}^{K}\cup\mathcal{V}_{u}^{U},
    \quad
    \mathcal{Y}(\mathcal{V}_{u}^{K})\subseteq\mathcal{C}^{K},
    \quad
    \mathcal{Y}(\mathcal{V}_{u}^{U})\subseteq\mathcal{C}_{t}^{U},
    \quad
    \mathcal{C}^{K}\cap\mathcal{C}_{t}^{U}=\emptyset .
\end{equation}

Let $\mathcal{K}_{t}$ denote the concept knowledge available to the learner
at stage $t$, initialized from the known concepts. Our goal goes beyond
merely separating known and unseen nodes. Given $\mathcal{G}$ and
$\mathcal{K}_{t}$, the learner should: (1) identify observations that cannot
be explained by existing concepts while preserving known-class recognition;
(2) infer coherent semantic concepts for the discovered unknown observations;
and (3) selectively incorporate reliable concepts to obtain an updated
knowledge state $\mathcal{K}_{t+1}$. Thus, open-world multimodal graph
learning is formulated as a progressive process from \emph{knowledge
insufficiency} to \emph{semantic discovery} and finally \emph{knowledge
evolution}.

\section{Related Work}
\label{sec:related}

\textbf{Open-World Graph Learning.}
Early open-world graph learning mainly focuses on rejecting nodes beyond the
known label space. OpenWGL~\citep{wu2020openwgl} models uncertainty for
unseen-node recognition, followed by studies that improve OOD detection through
predictive uncertainty~\citep{stadler2021graph}, neighborhood information, and
structure-aware decision boundaries. Recent works further move beyond binary
known--unknown separation: OpenIMA~\citep{wang2024openima} and
ORAL~\citep{jin2024beyond} explore novel class discovery, while OGA~\citep{wen2025llms}
extends the pipeline toward unknown recognition, semantic annotation, and model
updating with LLM assistance. Despite this progress, existing methods still
treat discovery, annotation, and updating as separate stages with different
criteria. MOVE instead unifies these decisions under a shared compression-based
knowledge evolution principle.

\textbf{Multimodal Graph Learning and Foundation Models.}
Multimodal graph learning has evolved from modality-specific encoding and graph
fusion toward foundation-model-driven semantic reasoning. Recent benchmarks
such as MAGB~\citep{yan2025graph} systematically study multimodal graph
understanding, while GraphGPT-O~\citep{fang2025graphgpto} and multimodal graph
foundation models such as PLANET~\citep{liu2026planet} improve cross-modal
alignment and transferable representation learning. However, these methods
mainly focus on improving the representation of existing semantic spaces and
assume that concepts are predefined. MOVE further considers when the current
semantic space becomes insufficient and verifies whether newly proposed concepts
are worthy of being incorporated into knowledge.

\section{Empirical Study}
\label{sec:empirical_study}

Before introducing MOVE, we empirically investigate the three observations behind our design: whether unknown concepts are jointly unexplained across modalities, whether semantic proposals are reliably grounded, and whether plausible concepts truly deserve knowledge expansion(see Appendix~\ref{app:empirical_details}).

\begin{figure*}[t]
    \centering
    \includegraphics[
        width=\textwidth
    ]{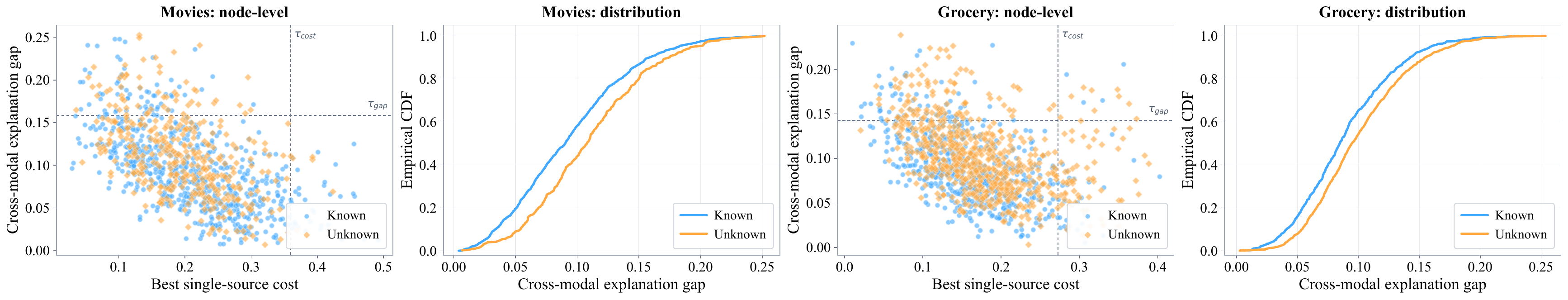}
    \caption{
\textbf{Cross-modal explanation gaps.} The panels compare explanation gaps between known and unknown nodes with their cumulative distributions.
}
    \label{fig:empirical_gap}
\end{figure*}
\begin{figure*}[t]
    \centering
    \includegraphics[
        width=\textwidth
    ]{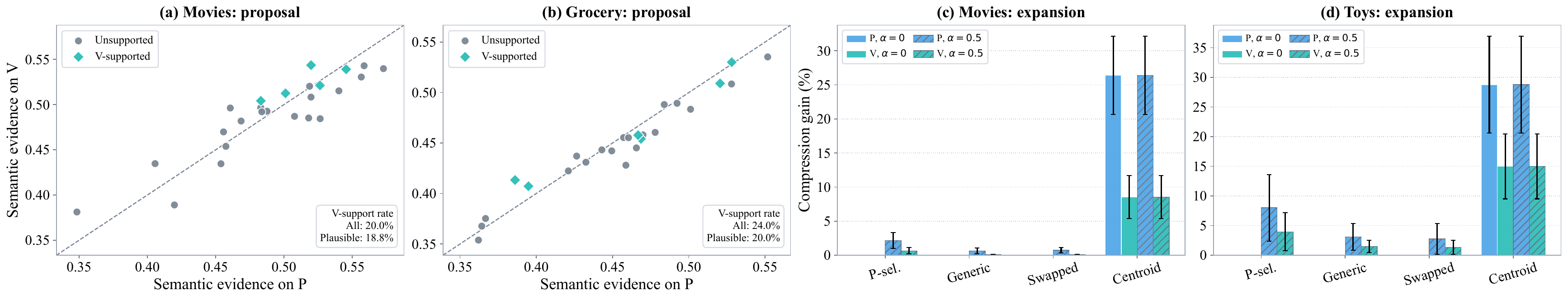}
    \caption{
\textbf{Semantic proposal ambiguity and class expansion.} (a--b) Candidate evidence on proposal set $P$ and validation set $V$; the in-panel statistics report multimodal support consistency on $V$. (c--d) Compression gains on proposal and validation evidence under different adaptation settings. Error bars denote standard errors across discovered regions.
}
    \label{fig:empirical_proposal_expansion}
\end{figure*}

\subsection{Multimodal Assignment Failure}
\label{sec:empirical_gap}

We first examine whether unknown concepts can be identified by individual modalities. For node $i$, we measure the gap between its best single-source match and the cost of jointly explaining visual, textual, and structural evidence:
\begin{equation}
G_i=
\min_k
\frac{1}{|\mathcal{R}|}
\sum_{r\in\mathcal{R}}
d(\mathbf{z}_i^r,\mathbf{c}_k^r)
-
\min_{r,k}
d(\mathbf{z}_i^r,\mathbf{c}_k^r),
\quad
\mathcal{R}=\{v,t,g\}.
\end{equation}
As shown in Fig.~\ref{fig:empirical_gap}, \textbf{13.8\%} of unknown nodes remain close to known concepts in at least one modality, but exhibit larger joint explanation gaps. This indicates that open-world novelty is better captured by multimodal explanation failure than isolated anomalies.

\subsection{Ambiguous Class Induction}
\label{sec:empirical_proposal}

We next investigate whether semantic proposals can be directly obtained from unresolved observations. Given unknown regions, we ask an MLLM to generate candidate concepts and evaluate their support from multimodal evidence. Fig.~\ref{fig:empirical_proposal_expansion}(a--b) shows that although multiple concepts can be linguistically plausible, only \textbf{19.4\%} receive consistent evidence support. This demonstrates that semantic generation alone is insufficient, motivating evidence-grounded hypothesis verification.

\subsection{Unreliable Class Expansion}
\label{sec:empirical_expansion}
Finally, we examine whether improving local fitting is sufficient for knowledge expansion. For a candidate concept $p$, we compare its compression gain on the proposal set and independent evidence:
\begin{equation}
G^{P/V}(p)
=
R(S_g^{P/V};\mathcal{K})
-
R(S_g^{P/V};\mathcal{K}\cup\{p\}).
\end{equation}
As shown in Fig.~\ref{fig:empirical_proposal_expansion}(c--d), generic concepts can reduce local fitting error by \textbf{1.9\%} due to increased capacity, but their gain decreases on independent evidence. Therefore, reliable knowledge evolution requires validating whether a concept provides transferable explanatory value rather than merely better fitting existing observations.

\section{Methodology}
\label{sec:methodology}

\subsection{Overview}
\label{sec:method_overview}

\begin{figure*}[t]
    \centering
    \includegraphics[width=\textwidth]{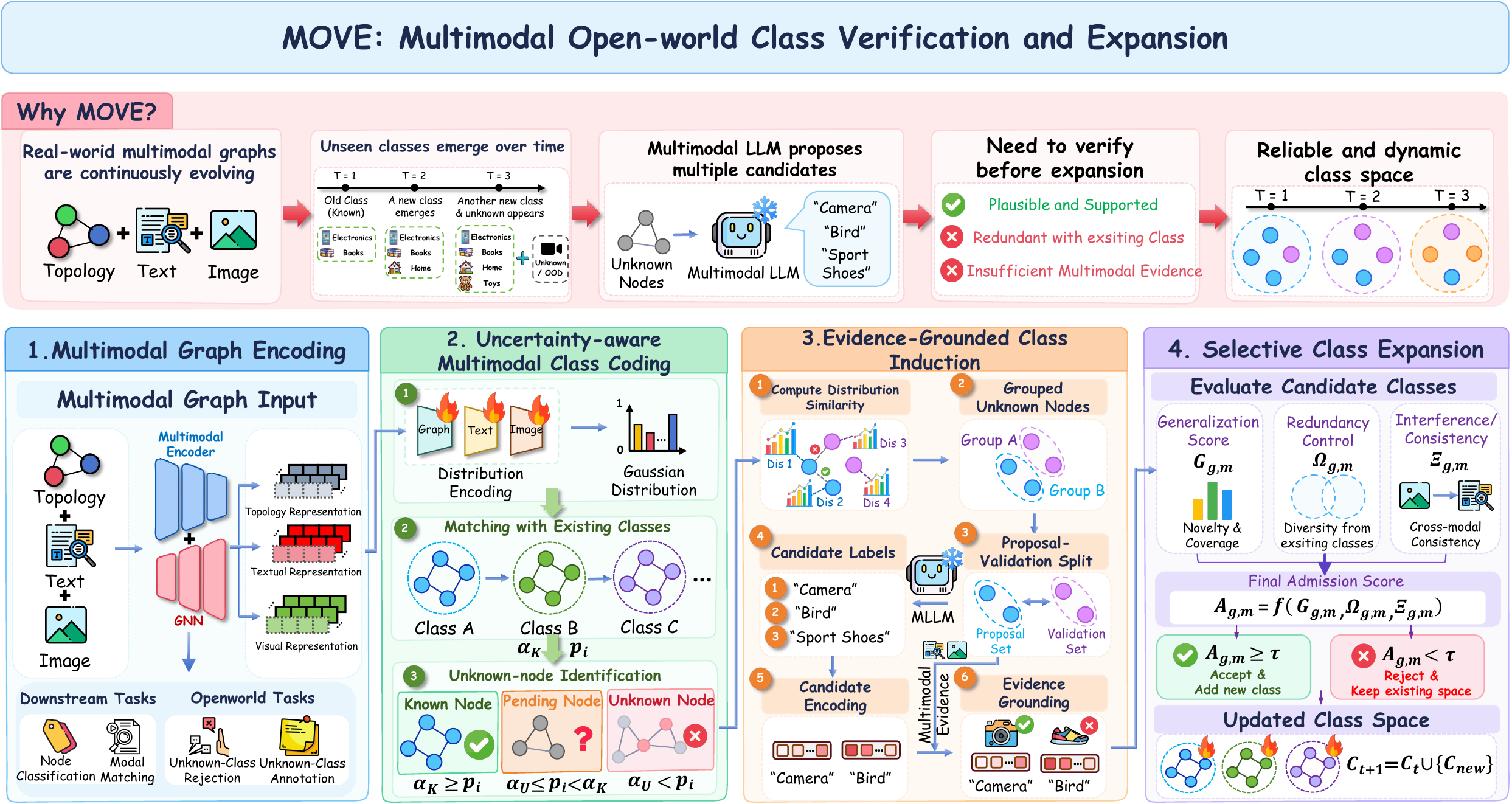}
    \caption{
    Overview of MOVE.
    }
    \label{fig:framework}
\end{figure*}

As shown in Fig.~\ref{fig:framework}, MOVE consists of three tightly coupled
stages. \textbf{Uncertainty-Aware Multimodal Class Coding} evaluates whether existing
classes can accommodate a node by jointly considering visual, textual, and
structural information. \textbf{Evidence-Grounded Class Induction} uses a
multimodal LLM to generate candidate classes for nodes that cannot be assigned
to existing classes and verifies them with multimodal evidence. Finally,
\textbf{Selective Class Expansion} determines whether a candidate class should
be incorporated by evaluating its ability to characterize previously unresolved
nodes while avoiding redundant categories and interference with existing
classes. Details are applied in App.\ref{app:method_details}.

\subsection{Uncertainty-Aware Multimodal Class Coding}
\label{sec:class_coding}

\textbf{Motivation.}
Obs. 1 in Sec.~\ref{sec:empirical_gap} shows that a node may
appear similar to known classes in an individual modality, while the joint
visual, textual, and structural information prevents it from being assigned to
any existing class. Therefore, MOVE identifies unknown nodes by evaluating
whether current classes can sufficiently represent their multimodal
characteristics rather than relying on isolated modality similarity.

\textbf{Distributional multimodal representation.}
To preserve uncertainty from heterogeneous modalities, MOVE represents each
node as modality-specific distributions instead of deterministic embeddings:
\begin{equation}
q_i^{r}(\mathbf{z})
=
\mathcal{N}
\left(
\boldsymbol{\mu}^{r}_{i},
\operatorname{diag}[(\boldsymbol{\sigma}^{r}_{i})^{2}]
\right),
\quad
(\boldsymbol{\mu}^{r}_{i},
\log\boldsymbol{\sigma}^{r}_{i})
=
\phi_r(\mathbf{x}^{r}_{i}),
\quad
r\in\{v,t,g\},
\label{eq:semantic_experts}
\end{equation}
where $\mathbf{x}^{v}_{i}$, $\mathbf{x}^{t}_{i}$, and $\mathbf{x}^{g}_{i}$
denote visual tokens, textual attributes, and graph-context representations,
respectively. Since different modalities may have different reliability, we
assign adaptive weights:
\[
\omega_i^{r}
\propto
\exp[-\mathcal{H}(q_i^{r})/\tau_{\omega}],
\quad
\sum_r\omega_i^{r}=1 .
\]

Given an existing class representation $p_k$, we evaluate how well class $k$
matches the multimodal information of node $i$:
\begin{equation}
E_{ik}
=
\sum_{r}
\omega_i^{r}
D_{\mathrm{SKL}}(q_i^{r}\Vert p_k)
+
\gamma
\sum_{r<s}
\omega_i^{r}\omega_i^{s}
D_{\mathrm{B}}(q_i^{r},q_i^{s}),
\quad
r,s\in\{v,t,g\}.
\label{eq:concept_energy}
\end{equation}
The first term measures the compatibility between a node and an existing class,
while the second term captures whether multimodal information can be jointly
aligned. Therefore, a node may remain close to known classes in one modality
but still fail to match any existing class when all modalities are considered.
Instead of assigning nodes according to the nearest class, MOVE evaluates the
overall assignment uncertainty over the existing class space:
\begin{equation}
\mathcal{R}_{i}(\mathcal{C}_{t})
=
-\tau_c
\log
\sum_{k=1}^{K_t}
\pi_k
\exp(-E_{ik}/\tau_c),
\quad
a_{ik}
=
\frac{
\pi_k\exp(-E_{ik}/\tau_c)}
{
\sum_l\pi_l\exp(-E_{il}/\tau_c)
}.
\label{eq:free_energy}
\end{equation}
where $\mathcal{C}_{t}$ denotes the current class space. A low coding cost
indicates that existing classes can represent the node, while a high cost
suggests that the node cannot be assigned to the current classes.
The class representations and modality encoders are optimized using known
training nodes:
\begin{equation}
\mathcal{L}_{\mathrm{code}}
=
-\frac{1}{|\mathcal{V}^{K}_{\mathrm{tr}}|}
\sum_{i\in\mathcal{V}^{K}_{\mathrm{tr}}}
\log
\frac{
\pi_{y_i}\exp(-E_{iy_i}/\tau_c)}
{
\sum_k\pi_k\exp(-E_{ik}/\tau_c)
}.
\label{eq:coding_training}
\end{equation}

\textbf{Unknown-node identification.}
Since coding costs vary across datasets, MOVE calibrates each node against
known-class references:
\begin{equation}
p_i=
\frac{
1+\sum_{r\in\mathcal{D}_{K}(c_i)}
\mathbb{I}[r\geq\mathcal{R}_{i}]
}
{
1+|\mathcal{D}_{K}(c_i)|
},
b_i=-\log_2p_i,
\operatorname{state}(i)=
\begin{cases}
\mathrm{known}, & p_i>\alpha_K,\\
\mathrm{pending}, & \alpha_U<p_i\leq\alpha_K,\\
\mathrm{unknown}, & p_i\leq\alpha_U .
\end{cases}
\label{eq:energy_calibration}
\end{equation}
Only nodes identified as unknown are used for discovering new classes, while
pending nodes provide additional supporting evidence.

\subsection{Evidence-Grounded Class Induction}
\label{sec:class_induction}

\textbf{Motivation.}
Obs. 2 in Sec.~\ref{sec:empirical_proposal} shows that nodes
outside the current class space do not directly reveal their missing class.
Although multimodal LLMs can generate plausible class descriptions, such
descriptions may not fully reflect visual, textual, and structural
characteristics. Therefore, MOVE treats LLM outputs as candidate class
hypotheses and verifies them using multimodal evidence.

\textbf{Multimodal evidence aggregation.}
For unknown nodes, MOVE first aggregates their multimodal information:
\begin{equation}
q_i(\mathbf{z})
=
\frac{1}{Z_i}
\prod_{r\in\{v,t,g\}}
[q_i^{r}(\mathbf{z})]^{\omega_i^{r}},
\quad
\{S_g\}_{g=1}^{G}
=
\operatorname{Cluster}
\left(
\{q_i\mid\operatorname{state}(i)=\mathrm{unknown}\};
D_{\mathrm{B}}
\right).
\label{eq:evidence_support}
\end{equation}

Each group is divided into proposal and validation subsets:
\[
S_g=S_g^{P}\cup S_g^{V},
\quad
S_g^{P}\cap S_g^{V}=\emptyset .
\]
The proposal subset is used for generating candidate classes, while the
validation subset is reserved for evaluating whether a candidate generalizes
beyond the observations used for generation.

\textbf{Candidate class generation and grounding.}
For each proposal group, multimodal observations are organized as an evidence
package $\mathcal{E}_g$. A multimodal LLM generates candidate class
descriptions $\mathcal{H}_g=\{h_{g,m}\}_{m=1}^{M}$, which are encoded into the
same semantic space as node representations as Gaussian class priors:
\begin{equation}
q^{h}_{g,m}
=
\mathcal{N}
(
\boldsymbol{\mu}^{h}_{g,m},
\boldsymbol{\Sigma}^{h}_{g,m}
),
\quad
h_{g,m}\in\mathcal{H}_g .
\label{eq:class_prior}
\end{equation}

To combine the semantic prior with observed evidence, MOVE first constructs a confidence-weighted, moment-matched group distribution $\bar q_g=\mathcal{N}(\bar{\boldsymbol{\mu}}_g, \bar{\boldsymbol{\Sigma}}_g)$, which preserves both within-node uncertainty and between-node scatter (Appendix~\ref{app:method_details}). Let $\bar{\boldsymbol{\Lambda}}_g=\bar{\boldsymbol{\Sigma}}_g^{-1}$. We then perform scale-preserving precision pooling:
\begin{equation}
\begin{aligned}
\boldsymbol{\Lambda}^{*}_{g,m}
&=
\frac{
\beta\boldsymbol{\Lambda}^{h}_{g,m}
+\bar{\boldsymbol{\Lambda}}_g
}{\beta+1},
\boldsymbol{\mu}^{*}_{g,m}
=
\left(
\beta\boldsymbol{\Lambda}^{h}_{g,m}
+\bar{\boldsymbol{\Lambda}}_g
\right)^{-1}
\left(
\beta\boldsymbol{\Lambda}^{h}_{g,m}
\boldsymbol{\mu}^{h}_{g,m}
+\bar{\boldsymbol{\Lambda}}_g
\bar{\boldsymbol{\mu}}_g
\right).
\end{aligned}
\label{eq:posterior_grounding}
\end{equation}
This normalization prevents the candidate covariance from shrinking merely as the proposal-group size increases.
The grounded candidate class is evaluated according to its ability to improve
class assignment:
\begin{equation}
u_{g,m}
=
\frac{1}{|S_g^{P}|}
\sum_{i\in S_g^{P}}
\left[
\mathcal{R}_{i}(\mathcal{C}_{t})
-
\mathcal{R}_{i}
(\mathcal{C}_{t}\cup\{p^{*}_{g,m}\})
\right],
\quad
\rho_{g,m}
=
\frac{\exp(u_{g,m}/\tau_h)}
{\sum_j\exp(u_{g,j}/\tau_h)} .
\label{eq:hypothesis_posterior}
\end{equation}

The LLM provides candidate class semantics, while multimodal evidence
determines whether the candidate is sufficiently supported.

\subsection{Selective Class Expansion}
\label{sec:class_expansion}

\textbf{Motivation.}
Obs. 3 in Sec.~\ref{sec:empirical_expansion} shows that a
candidate class that fits current unknown nodes may still introduce redundant
categories. Therefore, MOVE evaluates candidates on unseen evidence before
expanding the class space.

\textbf{Cross-validated class improvement.}
For candidate class $p^{*}_{g,m}$, MOVE evaluates its generalization ability on
the validation subset by measuring the reduction of class assignment cost:
\begin{equation}
G^{\mathrm{LCB}}_{g,m}
=
\frac{1}{|S_g^{V}|}
\sum_{i\in S_g^{V}}
\left[
\mathcal{R}_{i}(\mathcal{C}_{t})
-
\mathcal{R}_{i}
(\mathcal{C}_{t}\cup\{p^{*}_{g,m}\})
\right]
-
z_{1-\delta_{\mathrm{conf}}}
\frac{\widehat{\sigma}_{g,m}}
{\sqrt{|S_g^{V}|}} .
\label{eq:cross_fitted_gain}
\end{equation}

This validation process filters candidates that only fit the nodes used for
generation.

\textbf{Redundancy and interference control.}
To avoid unnecessary category expansion, MOVE considers class redundancy:
\begin{equation}
\begin{aligned}
R_{g,m}^{\mathrm{red}}
&=
\sum_{k=1}^{K_t}
\pi_k
\exp
\left(
-\frac{
D_{\mathrm{B}}(p^{*}_{g,m},p_k)
}
{\tau_{\Omega}}
\right),
C_{g,m}^{\mathrm{cmp}}
=
\frac{d_{\mathrm{eff}}}{2|S_g^{P}|}
\log(1+|S_g^{P}|),
\Omega_{g,m}
=
R_{g,m}^{\mathrm{red}}
+
\lambda_{\mathrm{cmp}}C_{g,m}^{\mathrm{cmp}}.
\end{aligned}
\label{eq:description_complexity}
\end{equation}

We further measure whether a new class changes assignments of existing nodes:
\begin{equation}
\Xi_{g,m}
=
\frac{1}{|\mathcal{B}_{\mathrm{old}}|}
\sum_{i\in\mathcal{B}_{\mathrm{old}}}
D_{\mathrm{JS}}
\left(
[\mathbf{a}^{-}_{i},0]
\Vert
\mathbf{a}^{+}_{i,g,m}
\right).
\label{eq:semantic_interference}
\end{equation}

The final admission score is:
\begin{equation}
\mathcal{A}_{g,m}
=
G^{\mathrm{LCB}}_{g,m}
-
\lambda_{\Omega}\Omega_{g,m}
-
\lambda_{\Xi}\Xi_{g,m},
\quad
m^{*}
=
\arg\max_m\mathcal{A}_{g,m}.
\label{eq:admission_score}
\end{equation}

The candidate class is incorporated only when it provides reliable improvement:
\begin{equation}
\mathcal{C}_{t+1}
=
\begin{cases}
\mathcal{C}_{t}
\cup
\{(\pi_{g,m^{*}},p^{*}_{g,m^{*}})\},
&
\mathcal{A}_{g,m^{*}}>0,\\
\mathcal{C}_{t},
&
\text{otherwise}.
\end{cases}
\label{eq:class_admission}
\end{equation}

\section{Experiments}
\label{sec:experiments}

\definecolor{ooddetshade}{HTML}{DDF5F5}
\definecolor{oodannshade}{HTML}{FFF4D6}
\definecolor{mmoodshade}{HTML}{E2EEF9}
\definecolor{openworldshade}{HTML}{EDE4F7}
\definecolor{oursshade}{HTML}{ECECEC}
\definecolor{ablationshade}{HTML}{F2F2F2}
\definecolor{lowerboundshade}{HTML}{F2F2F2}
\definecolor{upperboundshade}{HTML}{D9E2F3}

In this section, we conduct a wide range of experiments and aim to answer the following questions: \textbf{Q1: Effectiveness.} Compared with other SOTA baselines, can MOVE achieve superior performance on both UCR and UCA? \textbf{Q2: Interpretability.} How do the core MDL components contribute to the overall success of MOVE? \textbf{Q3: Robustness.} How sensitive is MOVE to key hyperparameters, and how does it perform under severe cross-modal semantic inconsistency? \textbf{Q4: Efficiency.} What is the computational efficiency of MOVE, particularly regarding the MDL calculation and LLM-based semantic proposals? \textbf{Q5: Convergence and Scalability.} How do MOVE and competing methods behave as training progresses, and how does MOVE scale with increasing graph and category sizes?

\subsection{Experimental Setup}
\label{subsec:experimental_setup}

We evaluate MOVE on four multimodal attributed graphs, including Toys, Grocery, Movies, and Reddit-S~\citep{yan2025graph}. We compare MOVE with representative open-world graph learning baselines, including OOD detection methods, OOD annotation methods, multimodal OOD methods and open-world methods. The evaluation covers the complete open-world pipeline, including unknown-class rejection (UCR), unknown-class annotation (UCA), downstream graph-level and modality-level task. Detailed descriptions are provided in Appendix~\ref{sec:experiment_details}.

\subsection{Performance Comparison (Q1)}
\label{subsec:performance_comparison}

\begin{table*}[t]
\centering

\begin{minipage}[t]{0.49\textwidth}
\centering
\caption{Known-class recognition in UCR.}
\label{tab:ucr_known_accuracy}

{\fontsize{5.5}{6.3}\selectfont
\renewcommand{\scriptsize}{\fontsize{4.2}{4.8}\selectfont}
\setlength{\tabcolsep}{0.5pt}
\renewcommand{\arraystretch}{1.05}
\begin{tabular*}{\linewidth}{@{}ll|@{\extracolsep{\fill}}cccc@{}}
\toprule
\multirow{2}{*}{\textbf{Category}} & \multirow{2}{*}{\textbf{Method}}
& \multicolumn{4}{c}{\textbf{Known-class Accuracy ($\uparrow$)}} \\
\cmidrule(lr){3-6}
& & \textbf{Toys}
& \textbf{Grocery}
& \textbf{Reddit-S}
& \textbf{Movies} \\
\midrule

\cellcolor{ooddetshade}
& \cellcolor{ooddetshade} TopoOOD
& \textcolor{orange}{85.39{\scriptsize$\pm$0.54}}
& 84.43{\scriptsize$\pm$0.54}
& 97.83{\scriptsize$\pm$0.10}
& \textcolor{orange}{59.74{\scriptsize$\pm$0.61}} \\

\cellcolor{ooddetshade}\multirow{-2}{*}{\shortstack{OOD\\Detection}}
& \cellcolor{ooddetshade} GNN-Safe
& \textcolor{blue}{85.45{\scriptsize$\pm$0.47}}
& 84.18{\scriptsize$\pm$0.52}
& \textcolor{blue}{97.95{\scriptsize$\pm$0.10}}
& \textcolor{blue}{59.79{\scriptsize$\pm$0.32}} \\

\midrule

\cellcolor{oodannshade}
& \cellcolor{oodannshade} ORAL
& 82.55{\scriptsize$\pm$0.80}
& 83.51{\scriptsize$\pm$0.31}
& 96.35{\scriptsize$\pm$0.14}
& 56.63{\scriptsize$\pm$0.74} \\

\cellcolor{oodannshade}\multirow{-2}{*}{\shortstack{OOD\\Annotation}}
& \cellcolor{oodannshade} OpenWGL
& 78.17{\scriptsize$\pm$0.64}
& 71.46{\scriptsize$\pm$1.55}
& 97.34{\scriptsize$\pm$0.70}
& 52.34{\scriptsize$\pm$0.82} \\

\midrule

\cellcolor{mmoodshade}
& \cellcolor{mmoodshade} OGA
& 85.02{\scriptsize$\pm$0.60}
& \textcolor{orange}{87.14{\scriptsize$\pm$1.06}}
& 95.50{\scriptsize$\pm$0.36}
& 59.40{\scriptsize$\pm$0.39} \\

\cellcolor{mmoodshade}\multirow{-2}{*}{MM-OOD}
& \cellcolor{mmoodshade} AHM
& 82.51{\scriptsize$\pm$0.73}
& 81.14{\scriptsize$\pm$0.90}
& 96.67{\scriptsize$\pm$0.62}
& 55.74{\scriptsize$\pm$0.71} \\

\midrule

\cellcolor{openworldshade}
& \cellcolor{openworldshade} CFC
& 63.89{\scriptsize$\pm$2.85}
& 51.62{\scriptsize$\pm$6.56}
& 79.07{\scriptsize$\pm$4.08}
& 38.88{\scriptsize$\pm$4.60} \\

\cellcolor{openworldshade}\multirow{-2}{*}{\shortstack{Open\\World}}
& \cellcolor{openworldshade} TopoNC
& 84.13{\scriptsize$\pm$0.21}
& \textcolor{blue}{89.64{\scriptsize$\pm$0.70}}
& \textcolor{orange}{97.85{\scriptsize$\pm$0.48}}
& 53.08{\scriptsize$\pm$1.03} \\

\midrule

\rowcolor{oursshade}
\textbf{Ours}
& \textbf{MOVE}
& \textcolor{red}{\textbf{86.65}{\scriptsize$\mathbf{\pm0.86}$}}
& \textcolor{red}{\textbf{90.12}{\scriptsize$\mathbf{\pm0.78}$}}
& \textcolor{red}{\textbf{98.02}{\scriptsize$\mathbf{\pm0.31}$}}
& \textcolor{red}{\textbf{61.31}{\scriptsize$\mathbf{\pm0.25}$}} \\

\bottomrule
\end{tabular*}
}
\end{minipage}
\hfill
\begin{minipage}[t]{0.49\textwidth}
\centering
\caption{Unknown-node rejection in UCR.}
\label{tab:ucr_unknown_rejection}

{\fontsize{5.5}{6.3}\selectfont
\setlength{\tabcolsep}{0.5pt}
\renewcommand{\arraystretch}{1.05}
\begin{tabular*}{\linewidth}{@{}ll|@{\extracolsep{\fill}}cccc@{}}
\toprule
\multirow{2}{*}{\textbf{Category}} & \multirow{2}{*}{\textbf{Method}}
& \multicolumn{4}{c}{\textbf{Coverage / Precision ($\uparrow/\uparrow$)}} \\
\cmidrule(lr){3-6}
& & \textbf{Toys}
& \textbf{Grocery}
& \textbf{Reddit-S}
& \textbf{Movies} \\
\midrule

\cellcolor{ooddetshade}
& \cellcolor{ooddetshade} TopoOOD
& 10.68/54.28
& 8.61/53.92
& 12.30/54.26
& 9.75/22.14 \\

\cellcolor{ooddetshade}\multirow{-2}{*}{\shortstack{OOD\\Detection}}
& \cellcolor{ooddetshade} GNN-Safe
& 62.79/\textcolor{orange}{85.74}
& 59.94/88.16
& \textcolor{orange}{90.88}/87.49
& 14.93/29.04 \\

\midrule

\cellcolor{oodannshade}
& \cellcolor{oodannshade} ORAL
& 30.67/74.23
& 35.99/82.15
& 41.80/77.10
& 15.05/29.52 \\

\cellcolor{oodannshade}\multirow{-2}{*}{\shortstack{OOD\\Annotation}}
& \cellcolor{oodannshade} OpenWGL
& \textcolor{red}{72.90}/\textcolor{red}{88.25}
& \textcolor{blue}{62.93}/88.83
& \textcolor{red}{95.49}/\textcolor{blue}{91.36}
& 22.39/\textcolor{red}{39.54} \\

\midrule

\cellcolor{mmoodshade}
& \cellcolor{mmoodshade} OGA
& 61.54/84.54
& 53.04/\textcolor{orange}{89.52}
& 81.71/84.32
& 17.82/32.46 \\

\cellcolor{mmoodshade}\multirow{-2}{*}{MM-OOD}
& \cellcolor{mmoodshade} AHM
& 36.33/76.86
& 30.26/79.64
& 89.74/83.65
& 16.26/31.09 \\

\midrule

\cellcolor{openworldshade}
& \cellcolor{openworldshade} CFC
& 60.70/80.58
& 47.98/84.09
& 89.57/88.79
& \textcolor{blue}{26.01}/28.93 \\

\cellcolor{openworldshade}\multirow{-2}{*}{\shortstack{Open\\World}}
& \cellcolor{openworldshade} TopoNC
& \textcolor{orange}{70.05}/83.65
& \textcolor{orange}{60.79}/\textcolor{blue}{89.83}
& 87.67/\textcolor{orange}{89.39}
& \textcolor{orange}{25.66}/\textcolor{orange}{33.66} \\

\midrule

\rowcolor{oursshade}
\textbf{Ours}
& \textbf{MOVE}
& \textcolor{blue}{\textbf{71.02}}/
  \textcolor{blue}{\textbf{87.62}}
& \textcolor{red}{\textbf{63.98}}/
  \textcolor{red}{\textbf{90.02}}
& \textcolor{blue}{\textbf{93.73}}/
  \textcolor{red}{\textbf{92.78}}
& \textcolor{red}{\textbf{38.81}}/
  \textcolor{blue}{\textbf{34.60}} \\

\bottomrule
\end{tabular*}
}
\end{minipage}

\end{table*}

\begin{table*}[t]
\centering
\caption{Semantic label quality in UCA.
Each item is expressed as generated-to-unknown (G-to-U) similarity ($\uparrow$) /
known-to-generated (K-to-G) similarity ($\downarrow$).}
\label{tab:uca_semantic_quality}

\resizebox{\textwidth}{!}{%
\begin{tabular}{ll|cccc}
\toprule
\multirow{2}{*}{\textbf{Category}} & \multirow{2}{*}{\textbf{Method}}
& \multicolumn{4}{c}{\textbf{Semantic Label Quality (G-to-U $\uparrow$ / K-to-G $\downarrow$)}} \\
\cmidrule(lr){3-6}
& & \textbf{Toys}
& \textbf{Grocery}
& \textbf{Reddit-S}
& \textbf{Movies} \\
\midrule

\cellcolor{oodannshade}
& \cellcolor{oodannshade} ORAL
& 53.17{\scriptsize$\pm$0.25} /
  \textcolor{orange}{58.54{\scriptsize$\pm$1.11}}
& 58.42{\scriptsize$\pm$0.90} /
  \textcolor{orange}{65.17{\scriptsize$\pm$0.78}}
& 79.40{\scriptsize$\pm$2.31} /
  83.74{\scriptsize$\pm$0.76}
& 60.96{\scriptsize$\pm$1.52} /
  73.71{\scriptsize$\pm$0.86} \\

\cellcolor{oodannshade}\multirow{-2}{*}{OOD Annotation}
& \cellcolor{oodannshade} OpenWGL
& 48.51{\scriptsize$\pm$3.31} /
  60.83{\scriptsize$\pm$1.58}
& 55.21{\scriptsize$\pm$0.93} /
  67.25{\scriptsize$\pm$1.58}
& 78.47{\scriptsize$\pm$3.83} /
  83.81{\scriptsize$\pm$0.32}
& 60.88{\scriptsize$\pm$0.31} /
  \textcolor{orange}{73.22{\scriptsize$\pm$0.54}} \\

\midrule

\cellcolor{mmoodshade}
& \cellcolor{mmoodshade} OGA
& 53.86{\scriptsize$\pm$0.62} /
  \textcolor{blue}{58.22{\scriptsize$\pm$0.86}}
& 58.78{\scriptsize$\pm$0.92} /
  \textcolor{blue}{63.73{\scriptsize$\pm$0.66}}
& \textcolor{orange}{79.46{\scriptsize$\pm$1.86}} /
  85.97{\scriptsize$\pm$1.08}
& 60.01{\scriptsize$\pm$0.38} /
  \textcolor{blue}{72.80{\scriptsize$\pm$0.58}} \\

\cellcolor{mmoodshade}\multirow{-2}{*}{MM-OOD}
& \cellcolor{mmoodshade} AHM
& 53.53{\scriptsize$\pm$2.70} /
  60.64{\scriptsize$\pm$1.38}
& 56.09{\scriptsize$\pm$1.08} /
  67.10{\scriptsize$\pm$0.79}
& \textcolor{blue}{81.29{\scriptsize$\pm$0.73}} /
  \textcolor{blue}{82.33{\scriptsize$\pm$1.01}}
& 62.21{\scriptsize$\pm$0.58} /
  73.29{\scriptsize$\pm$0.52} \\

\midrule

\cellcolor{openworldshade}
& \cellcolor{openworldshade} CFC
& \textcolor{blue}{57.00{\scriptsize$\pm$7.33}} /
  71.88{\scriptsize$\pm$12.48}
& \textcolor{blue}{60.64{\scriptsize$\pm$7.22}} /
  76.80{\scriptsize$\pm$11.70}
& 77.30{\scriptsize$\pm$8.55} /
  84.07{\scriptsize$\pm$6.95}
& \textcolor{orange}{64.50{\scriptsize$\pm$5.44}} /
  82.32{\scriptsize$\pm$8.74} \\

\cellcolor{openworldshade}\multirow{-2}{*}{\shortstack{Open\\World}}
& \cellcolor{openworldshade} TopoNC
& \textcolor{orange}{56.29{\scriptsize$\pm$6.74}} /
  73.13{\scriptsize$\pm$12.00}
& \textcolor{orange}{59.88{\scriptsize$\pm$5.87}} /
  77.75{\scriptsize$\pm$12.86}
& 77.31{\scriptsize$\pm$8.61} /
  \textcolor{orange}{83.08{\scriptsize$\pm$6.98}}
& \textcolor{blue}{65.22{\scriptsize$\pm$5.16}} /
  83.01{\scriptsize$\pm$10.08} \\

\midrule

\rowcolor{oursshade}
\textbf{Ours}
& \textbf{MOVE}
& \textcolor{red}{\textbf{64.39}{\scriptsize$\mathbf{\pm 3.11}$}} /
  \textcolor{red}{\textbf{53.84}{\scriptsize$\mathbf{\pm 5.18}$}}
& \textcolor{red}{\textbf{65.78}{\scriptsize$\mathbf{\pm 2.70}$}} /
  \textcolor{red}{\textbf{53.76}{\scriptsize$\mathbf{\pm 2.17}$}}
& \textcolor{red}{\textbf{86.09}{\scriptsize$\mathbf{\pm 0.78}$}} /
  \textcolor{red}{\textbf{81.86}{\scriptsize$\mathbf{\pm 0.48}$}}
& \textcolor{red}{\textbf{78.36}{\scriptsize$\mathbf{\pm 2.32}$}} /
  \textcolor{red}{\textbf{72.47}{\scriptsize$\mathbf{\pm 3.13}$}} \\

\bottomrule
\end{tabular}%
}
\end{table*}


\begin{table*}[t]
\centering
\caption{
Downstream performance after UCA on node classification (NC) and zero-shot modality matching (MM) ($\%$, $\uparrow$).
For NC, the lower bound uses no unknown-label annotation and the upper bound uses ground-truth unknown labels.
}
\label{tab:downstream_tasks}

\setlength{\tabcolsep}{3pt}
\resizebox{\textwidth}{!}{%
\begin{tabular}{ll|cc|cc|cc|cc}
\toprule
\multirow{2}{*}{\textbf{Category}} & \multirow{2}{*}{\textbf{Method}}
& \multicolumn{2}{c|}{\textbf{Toys}}
& \multicolumn{2}{c|}{\textbf{Grocery}}
& \multicolumn{2}{c|}{\textbf{Reddit-S}}
& \multicolumn{2}{c}{\textbf{Movies}} \\
\cmidrule(lr){3-4}
\cmidrule(lr){5-6}
\cmidrule(lr){7-8}
\cmidrule(lr){9-10}
& & \textbf{NC (Acc.)} & \textbf{MM}
& \textbf{NC (Acc.)} & \textbf{MM}
& \textbf{NC (Acc.)} & \textbf{MM}
& \textbf{NC (Acc.)} & \textbf{MM} \\
\midrule

\rowcolor{lowerboundshade}
Reference
& Lower bound
& 39.44{\scriptsize$\pm$0.55} & -
& 32.38{\scriptsize$\pm$0.67} & -
& 39.49{\scriptsize$\pm$1.11} & -
& 39.57{\scriptsize$\pm$0.67} & - \\

\midrule

\cellcolor{oodannshade}
& \cellcolor{oodannshade} ORAL
& \textcolor{orange}{41.68{\scriptsize$\pm$1.98}}
& \textcolor{red}{79.52{\scriptsize$\pm$2.69}}
& 32.05{\scriptsize$\pm$0.79}
& 76.74{\scriptsize$\pm$2.89}
& \textcolor{blue}{53.57{\scriptsize$\pm$2.94}}
& 68.16{\scriptsize$\pm$6.32}
& 41.70{\scriptsize$\pm$0.68}
& \textcolor{blue}{49.25{\scriptsize$\pm$9.74}} \\

\cellcolor{oodannshade}\multirow{-2}{*}{\shortstack{OOD\\Annotation}}
& \cellcolor{oodannshade} OpenWGL
& 39.70{\scriptsize$\pm$0.72}
& 11.32{\scriptsize$\pm$2.69}
& 29.85{\scriptsize$\pm$0.26}
& 22.99{\scriptsize$\pm$6.84}
& 43.09{\scriptsize$\pm$2.87}
& 55.09{\scriptsize$\pm$1.44}
& 42.26{\scriptsize$\pm$0.51}
& 17.93{\scriptsize$\pm$2.17} \\

\midrule

\cellcolor{mmoodshade}
& \cellcolor{mmoodshade} OGA
& \textcolor{blue}{42.57{\scriptsize$\pm$0.55}}
& 72.70{\scriptsize$\pm$3.42}
& \textcolor{blue}{36.74{\scriptsize$\pm$1.21}}
& \textcolor{blue}{83.76{\scriptsize$\pm$4.38}}
& 47.38{\scriptsize$\pm$4.33}
& \textcolor{blue}{82.61{\scriptsize$\pm$1.35}}
& 41.81{\scriptsize$\pm$0.60}
& 38.77{\scriptsize$\pm$9.90} \\

\cellcolor{mmoodshade}\multirow{-2}{*}{MM-OOD}
& \cellcolor{mmoodshade} AHM
& 38.58{\scriptsize$\pm$0.75}
& 66.88{\scriptsize$\pm$6.41}
& 29.57{\scriptsize$\pm$0.67}
& 32.26{\scriptsize$\pm$1.77}
& \textcolor{orange}{50.24{\scriptsize$\pm$1.35}}
& 55.15{\scriptsize$\pm$8.36}
& 42.16{\scriptsize$\pm$0.55}
& 29.82{\scriptsize$\pm$4.05} \\

\midrule

\cellcolor{openworldshade}
& \cellcolor{openworldshade} CFC
& 36.18{\scriptsize$\pm$2.15}
& \textcolor{orange}{74.12{\scriptsize$\pm$1.23}}
& \textcolor{orange}{36.36{\scriptsize$\pm$3.20}}
& 81.02{\scriptsize$\pm$3.44}
& 48.04{\scriptsize$\pm$0.84}
& \textcolor{orange}{81.86{\scriptsize$\pm$0.71}}
& \textcolor{orange}{45.46{\scriptsize$\pm$0.82}}
& 44.58{\scriptsize$\pm$0.97} \\

\cellcolor{openworldshade}\multirow{-2}{*}{\shortstack{Open\\World}}
& \cellcolor{openworldshade} TopoNC
& 31.24{\scriptsize$\pm$2.25}
& 73.67{\scriptsize$\pm$0.67}
& 33.72{\scriptsize$\pm$1.02}
& \textcolor{orange}{82.97{\scriptsize$\pm$0.60}}
& 48.48{\scriptsize$\pm$0.74}
& 81.83{\scriptsize$\pm$1.03}
& \textcolor{blue}{45.81{\scriptsize$\pm$1.04}}
& \textcolor{orange}{46.27{\scriptsize$\pm$3.66}} \\

\midrule

\rowcolor{oursshade}
\textbf{Ours}
& \textbf{MOVE}
& \textcolor{red}{\textbf{46.65}{\scriptsize$\mathbf{\pm1.74}$}}
& \textcolor{blue}{\textbf{76.69}{\scriptsize$\mathbf{\pm1.64}$}}
& \textcolor{red}{\textbf{37.60}{\scriptsize$\mathbf{\pm1.40}$}}
& \textcolor{red}{\textbf{84.87}{\scriptsize$\mathbf{\pm3.25}$}}
& \textcolor{red}{\textbf{57.27}{\scriptsize$\mathbf{\pm3.02}$}}
& \textcolor{red}{\textbf{86.54}{\scriptsize$\mathbf{\pm3.93}$}}
& \textcolor{red}{\textbf{46.80}{\scriptsize$\mathbf{\pm0.72}$}}
& \textcolor{red}{\textbf{50.95}{\scriptsize$\mathbf{\pm2.46}$}} \\

\midrule

\rowcolor{upperboundshade}
Reference
& Upper bound
& 72.82{\scriptsize$\pm$0.36} & -
& 82.13{\scriptsize$\pm$0.66} & -
& 94.60{\scriptsize$\pm$0.41} & -
& 53.90{\scriptsize$\pm$0.82} & - \\

\bottomrule
\end{tabular}%
}
\end{table*}

\textbf{UCR Performance.}
Tables~\ref{tab:ucr_known_accuracy} and~\ref{tab:ucr_unknown_rejection} report known-class recognition and unknown-node rejection, respectively. In all tables, red, blue, and orange values indicate the best, second-best, and third-best results, respectively. MOVE achieves the best known-class Accuracy on all four datasets, outperforming the strongest baseline by up to 2.98 percentage points. For unknown-node rejection, it ranks among the top two methods in both Coverage and Precision across all datasets. These results demonstrate that MOVE achieves a favorable balance between known-class recognition and unknown-class rejection.

\textbf{UCA Performance.}
For semantic label quality, ``K'', ``U'', and ``G'' denote known, held-out unknown, and generated labels, respectively. As shown in Table~\ref{tab:uca_semantic_quality}, MOVE consistently achieves the highest G-to-U similarity and the lowest K-to-G similarity across all four datasets. This indicates that its generated concepts accurately capture unknown-class semantics while remaining distinguishable from the known label space.

\textbf{Performance on Downstream Tasks.}
As shown in Table~\ref{tab:downstream_tasks}, MOVE achieves the best node classification performance among the compared annotation methods on all four datasets and consistently improves over the unannotated lower bound. For modality matching, it ranks first on three datasets and second on Toys. Together, these results show that the generated concepts provide useful supervision for both node classification and cross-modal alignment.

\subsection{Ablation Analysis (Q2)}
\label{subsec:ablation}

\begin{table*}[t]
\centering
\caption{
Ablation study of MOVE on four datasets. UCR reports Coverage/Precision, and NC reports post-annotation GCN node-classification Accuracy (\%).
}
\label{tab:move_ablation}
\resizebox{\textwidth}{!}{%
\begin{tabular}{l|cc|cc|cc|cc}
\toprule
\multirow{2}{*}{\textbf{Variants}}
& \multicolumn{2}{c|}{\textbf{Toys}}
& \multicolumn{2}{c|}{\textbf{Grocery}}
& \multicolumn{2}{c|}{\textbf{Movies}}
& \multicolumn{2}{c}{\textbf{Reddit-S}} \\
\cmidrule(lr){2-3}
\cmidrule(lr){4-5}
\cmidrule(lr){6-7}
\cmidrule(lr){8-9}
& UCR (Cov./Prec.) & NC (Acc.)
& UCR (Cov./Prec.) & NC (Acc.)
& UCR (Cov./Prec.) & NC (Acc.)
& UCR (Cov./Prec.) & NC (Acc.) \\
\midrule
\cellcolor{ablationshade} w/o CDM
& 70.65 / 80.68 & 44.66
& 62.34 / 88.73 & 32.49
& 36.89 / 29.79 & 45.59
& 92.31 / 91.64 & 52.45 \\

\cellcolor{ablationshade} w/o HCV
& 69.65 / 80.23 & 40.01
& 62.39 / 88.85 & 32.32
& 35.76 / 30.33 & 44.29
& 91.74 / 92.22 & 38.99 \\

\cellcolor{ablationshade} w/o GCE
& 68.39 / 80.55 & 44.82
& 62.25 / 88.78 & 36.77
& 34.48 / 30.42 & 45.86
& 86.23 / 92.42 & 53.12 \\

\cellcolor{ablationshade} w/o CSU
& -/- & 40.01
& -/- & 32.32
& -/- & 44.84
& -/- & 39.03 \\

\midrule
\rowcolor{oursshade}
\textbf{MOVE (full)}
& \textbf{71.02 / 87.62} & \textbf{46.65}
& \textbf{63.98 / 90.02} & \textbf{37.60}
& \textbf{38.81 / 34.60} & \textbf{46.80}
& \textbf{93.73 / 92.78} & \textbf{57.27} \\

\bottomrule
\end{tabular}%
}
\end{table*}

We evaluate four functional components of MOVE: Cross-Modal Disagreement Modeling (CDM), Held-out Candidate Validation (HCV), Graph-Context Encoding (GCE), and Class-Space Update (CSU). As shown in Table~\ref{tab:move_ablation}, CDM and GCE mainly benefit unknown-class rejection, whereas HCV and CSU are especially important for downstream learning. In particular, removing HCV or CSU reduces node-classification Accuracy on Reddit-S by more than 18 percentage points. Detailed ablation settings are provided in Appendix~\ref{sec:experiment_details}.

\subsection{Robustness Analysis (Q3)}
\label{subsec:robustness_analysis}

\begin{figure*}[t]
\centering

\begin{minipage}[t]{0.235\textwidth}
    \centering
    \includegraphics[width=0.98\linewidth]
    {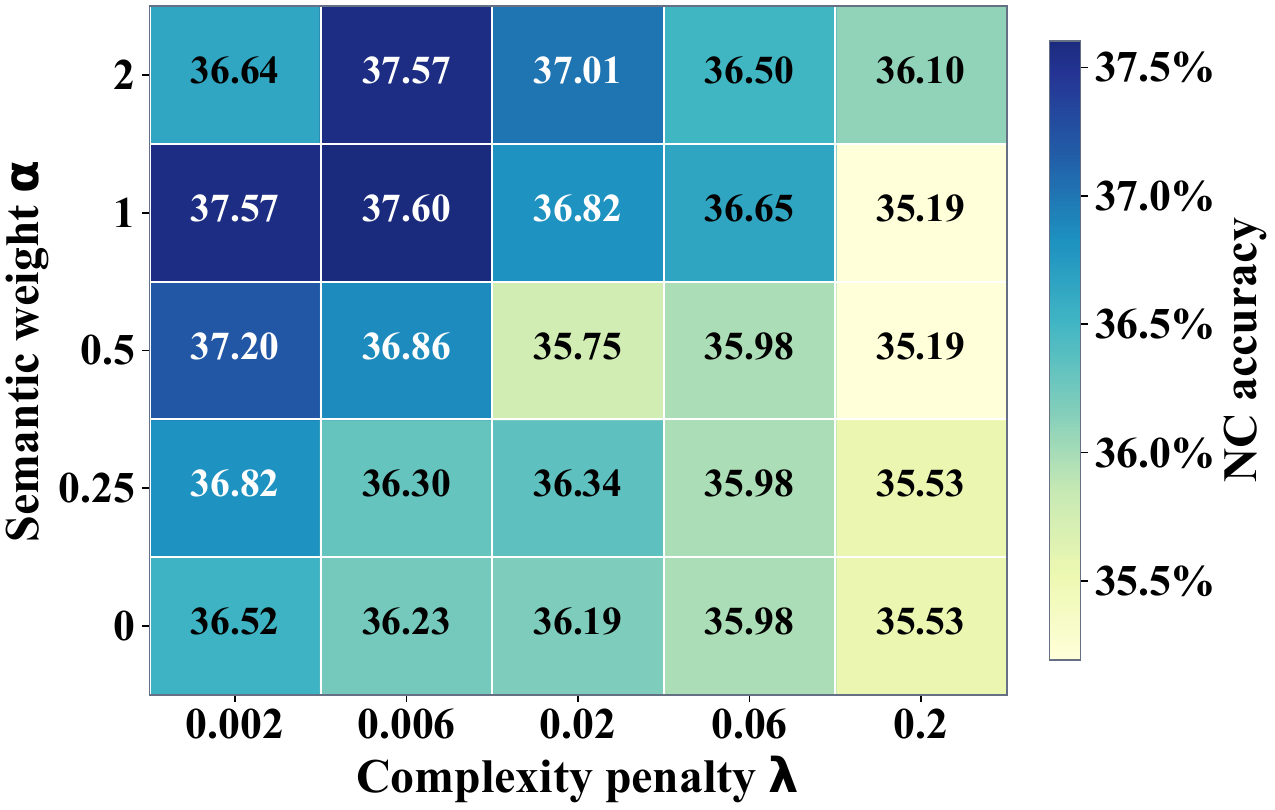}\\[-2pt]
    \small (a) Grocery: $\lambda$--$\alpha$.
\end{minipage}
\hfill
\begin{minipage}[t]{0.235\textwidth}
    \centering
    \includegraphics[width=0.98\linewidth]
    {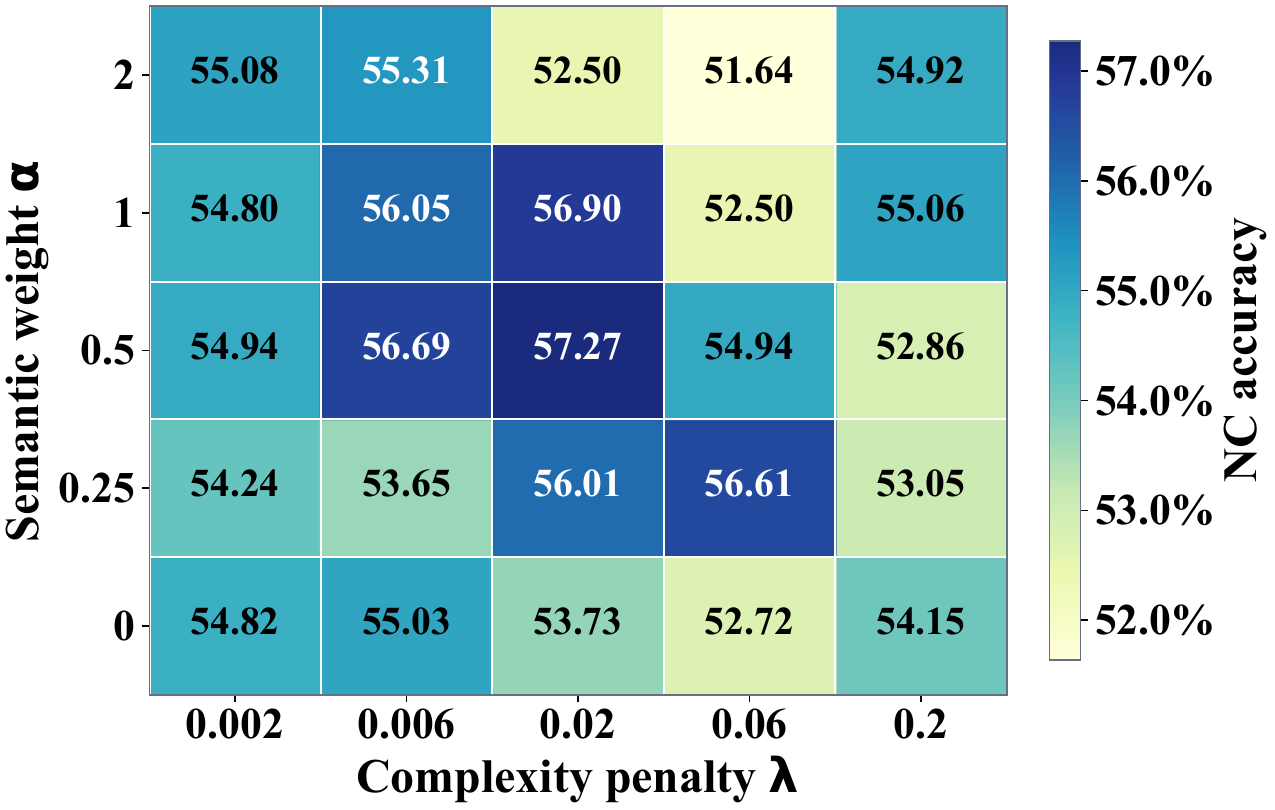}\\[-2pt]
    \small (b) Reddit-S: $\lambda$--$\alpha$.
\end{minipage}
\hfill
\begin{minipage}[t]{0.235\textwidth}
    \centering
    \includegraphics[width=0.98\linewidth]
    {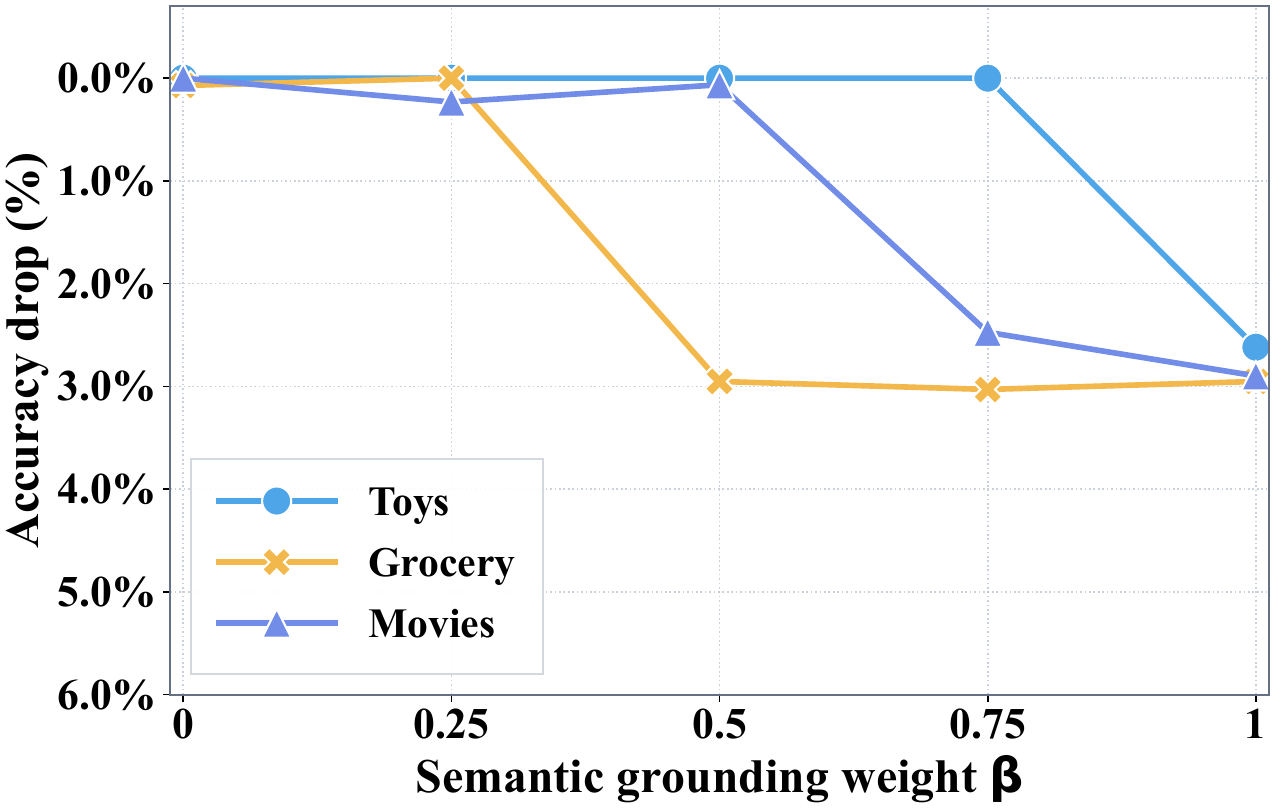}\\[-2pt]
    \small (c) Sensitivity to $\beta$.
\end{minipage}
\hfill
\begin{minipage}[t]{0.235\textwidth}
    \centering
    \includegraphics[width=0.98\linewidth]
    {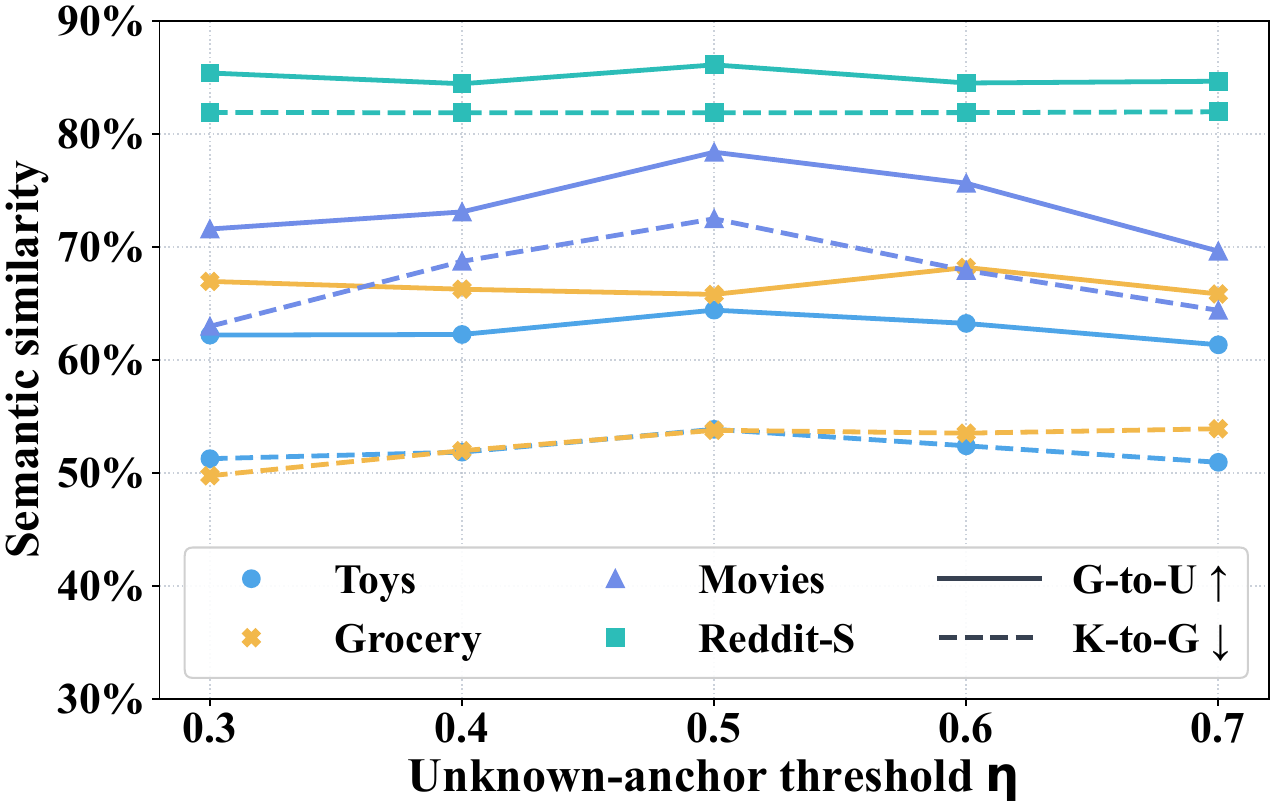}\\[-2pt]
    \small (d) Sensitivity to $\eta$.
\end{minipage}

\caption{
\textbf{Hyperparameter robustness of MOVE.}
(a)--(b) Post-annotation NC accuracy under different knowledge-complexity penalty $\lambda$ and conditional semantic-complexity weight $\alpha$.
(c) Relative drop in post-annotation NC accuracy under changes to the semantic grounding weight $\beta$.
(d) Semantic label quality under changes to the unknown-anchor evidence threshold
$\eta=\texttt{min\_unknown\_anchor\_fraction}$; solid and dashed lines denote G-to-U and K-to-G similarity, respectively.
}
\label{fig:robustness_hyperparameters}
\end{figure*}

\begin{figure*}[t]
\centering

\begin{minipage}[t]{0.49\textwidth}
    \centering
    \begin{minipage}[t]{0.485\linewidth}
        \centering
        \includegraphics[width=0.98\linewidth]
        {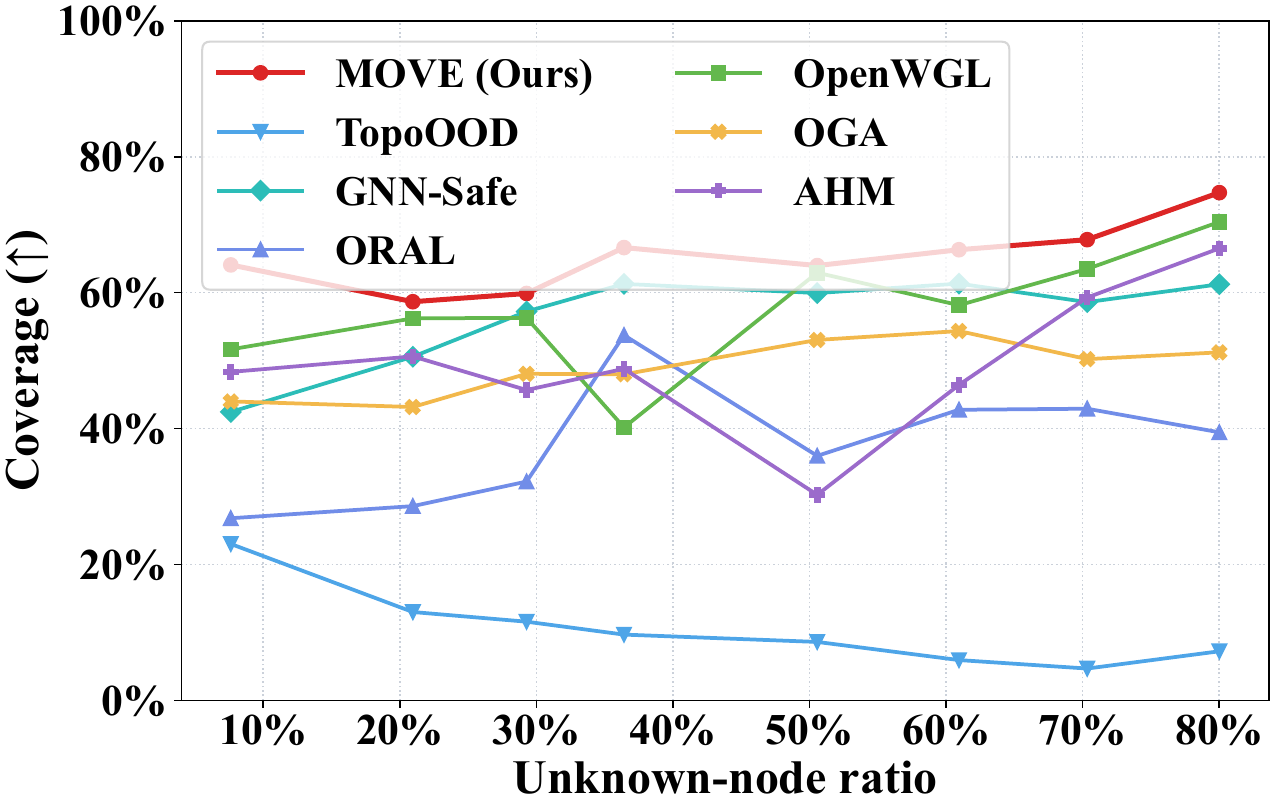}\\[-2pt]
        \small (a) Coverage.
    \end{minipage}
    \hfill
    \begin{minipage}[t]{0.485\linewidth}
        \centering
        \includegraphics[width=0.98\linewidth]
        {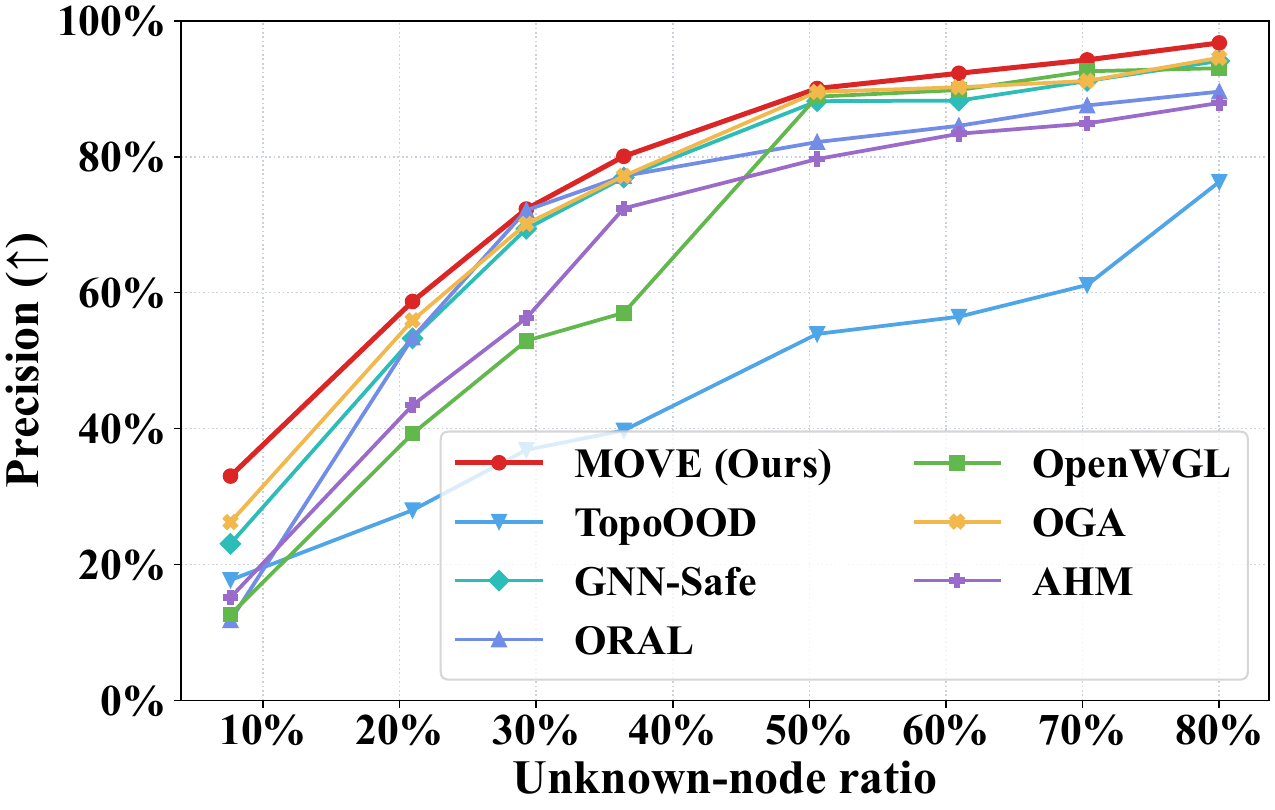}\\[-2pt]
        \small (b) Precision.
    \end{minipage}

    \caption{
    \textbf{Robustness to increasing graph openness.}
    Unknown-node Coverage and Precision on Grocery under increasing actual unknown-node ratios.
    }
    \label{fig:robustness_openness}
\end{minipage}
\hfill
\begin{minipage}[t]{0.49\textwidth}
    \centering
    \begin{minipage}[t]{0.485\linewidth}
        \centering
        \includegraphics[width=0.98\linewidth]{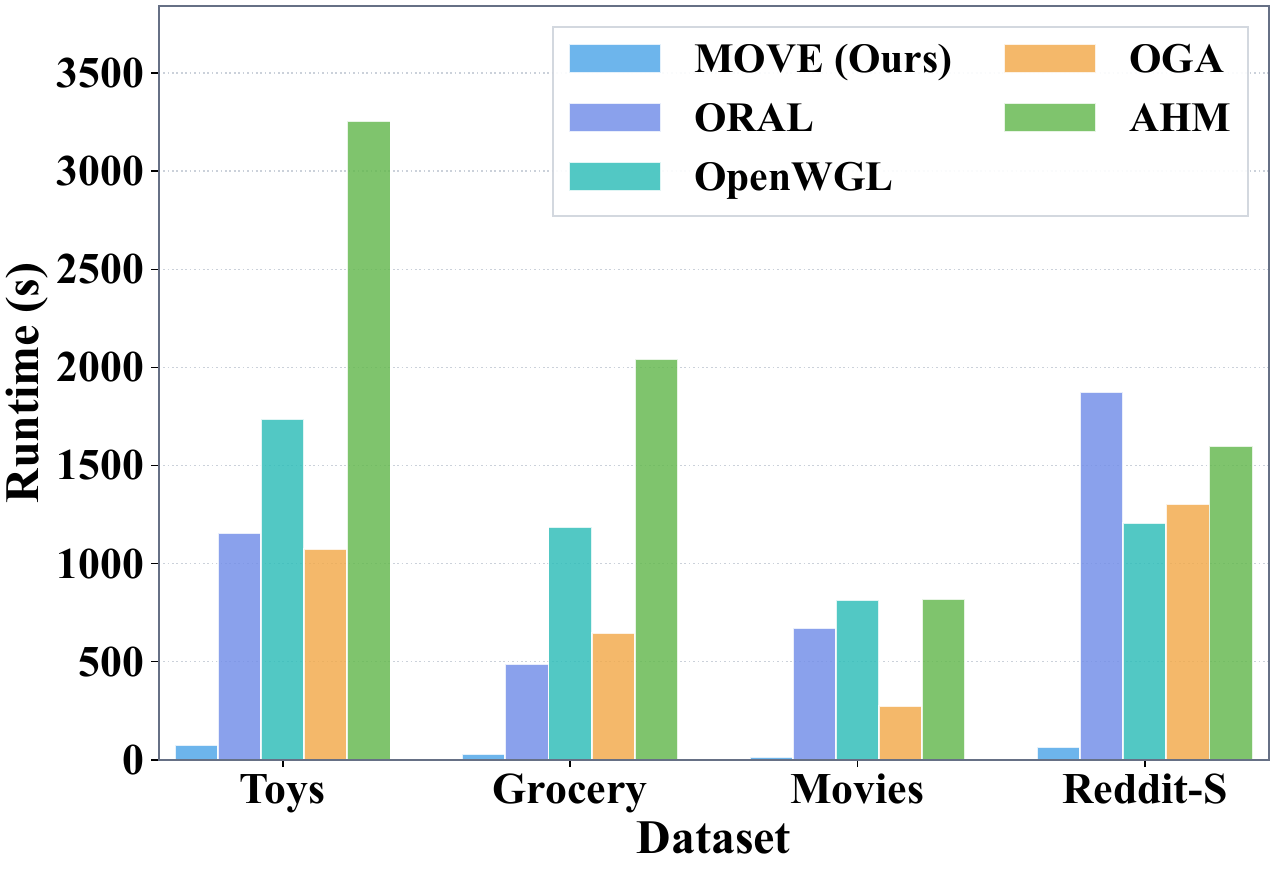}
        \small (a) Runtime.
    \end{minipage}
    \hfill
    \begin{minipage}[t]{0.485\linewidth}
        \centering
        \includegraphics[width=0.98\linewidth]{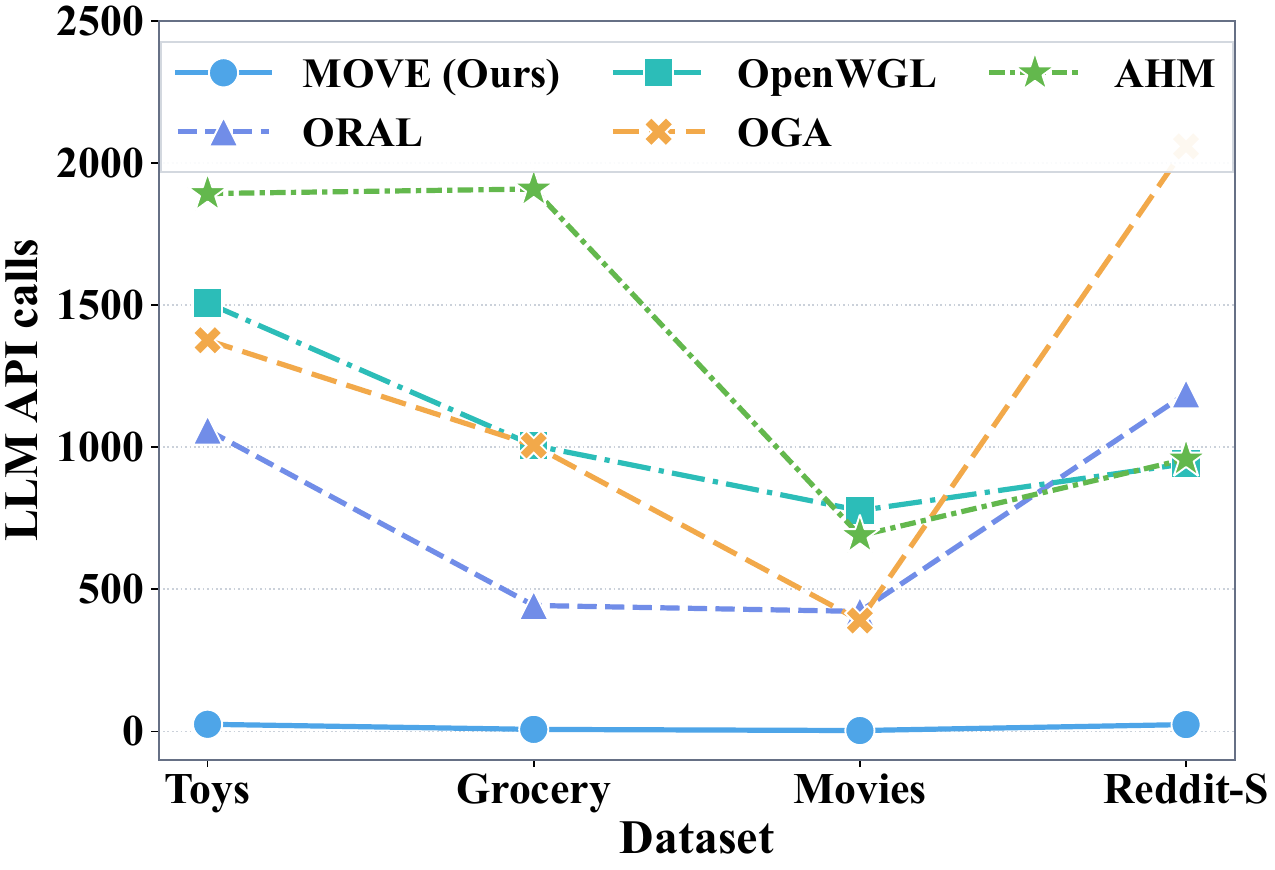}
        \small (b) LLM API calls.
    \end{minipage}

    \caption{
    \textbf{Efficiency and computational cost.}
    Runtime and actual LLM API calls are reported separately for MOVE and four annotation-capable baselines.
    }
    \label{fig:efficiency}
\end{minipage}
\end{figure*}

To answer Q3, we evaluate robustness from three perspectives: sensitivity to complexity controls, key hyperparameters, and increasing unknown-node ratios.

\textbf{Hyperparameter Sensitivity.}
Figures~\ref{fig:robustness_hyperparameters}(a)--(d) show that MOVE remains stable across different settings of complexity penalties $(\lambda,\alpha)$, semantic grounding weight $\beta$, and unknown-anchor threshold $\eta$. Moderate regularization achieves a balance between accepting useful new classes and avoiding redundant expansion. The effect of $\beta$ is dataset dependent, while the default $\eta=0.5$ provides a robust trade-off between G-to-U similarity and K-to-G similarity.

\textbf{Robustness to Increasing Openness.}
We increase the unknown-node ratio from 10\% to 80\% on Grocery and compare MOVE with representative open-world baselines under identical settings. As shown in Fig.~\ref{fig:robustness_openness}, MOVE consistently achieves strong Coverage and Precision across different openness levels, with stable unknown-node rejection performance even when known-class supervision becomes increasingly limited. These results demonstrate the robustness of MOVE under challenging open-world scenarios.

\subsection{Efficiency Comparison (Q4)}
\label{subsec:efficiency_comparison}

To answer Q4, Fig.~\ref{fig:efficiency} compares the annotation-stage efficiency of MOVE with annotation-capable baselines in terms of wall-clock time and LLM API calls. MOVE achieves the lowest runtime across all datasets, being $14.1\times$--$18.1\times$ faster than the fastest baseline while reducing LLM calls from hundreds or thousands to only 3--25. These improvements result from the MDL/UCR gate and compact candidate grouping, which avoid unnecessary LLM queries by validating only coherent candidate regions before concept expansion.

\subsection{Convergence and Scalability Analysis (Q5)}
\label{subsec:convergence_scalability}

\begin{figure*}[t]
\centering

\begin{minipage}[t]{0.49\textwidth}
\centering

\begin{minipage}[t]{0.485\linewidth}
    \centering
    \includegraphics[width=0.98\linewidth]
    {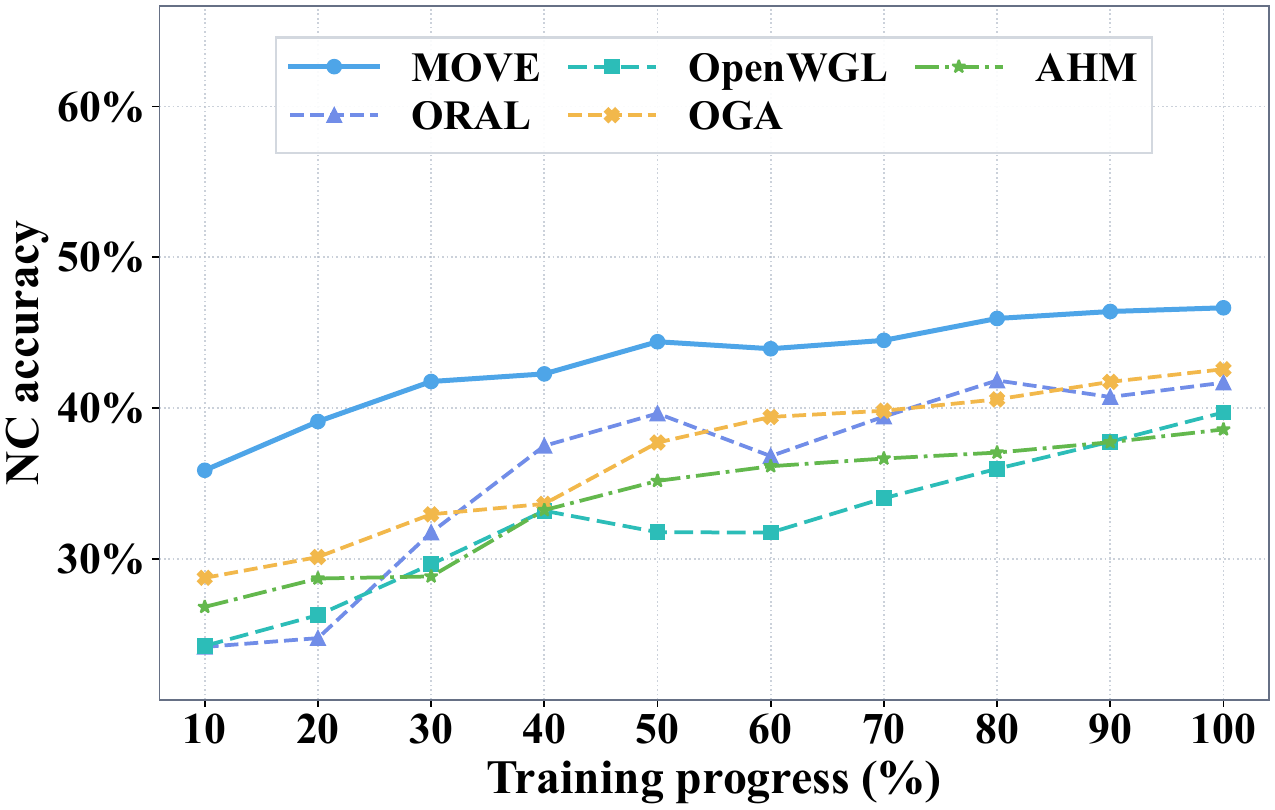}\\[-2pt]
    \small (a) Toys.
\end{minipage}
\hfill
\begin{minipage}[t]{0.485\linewidth}
    \centering
    \includegraphics[width=0.98\linewidth]
    {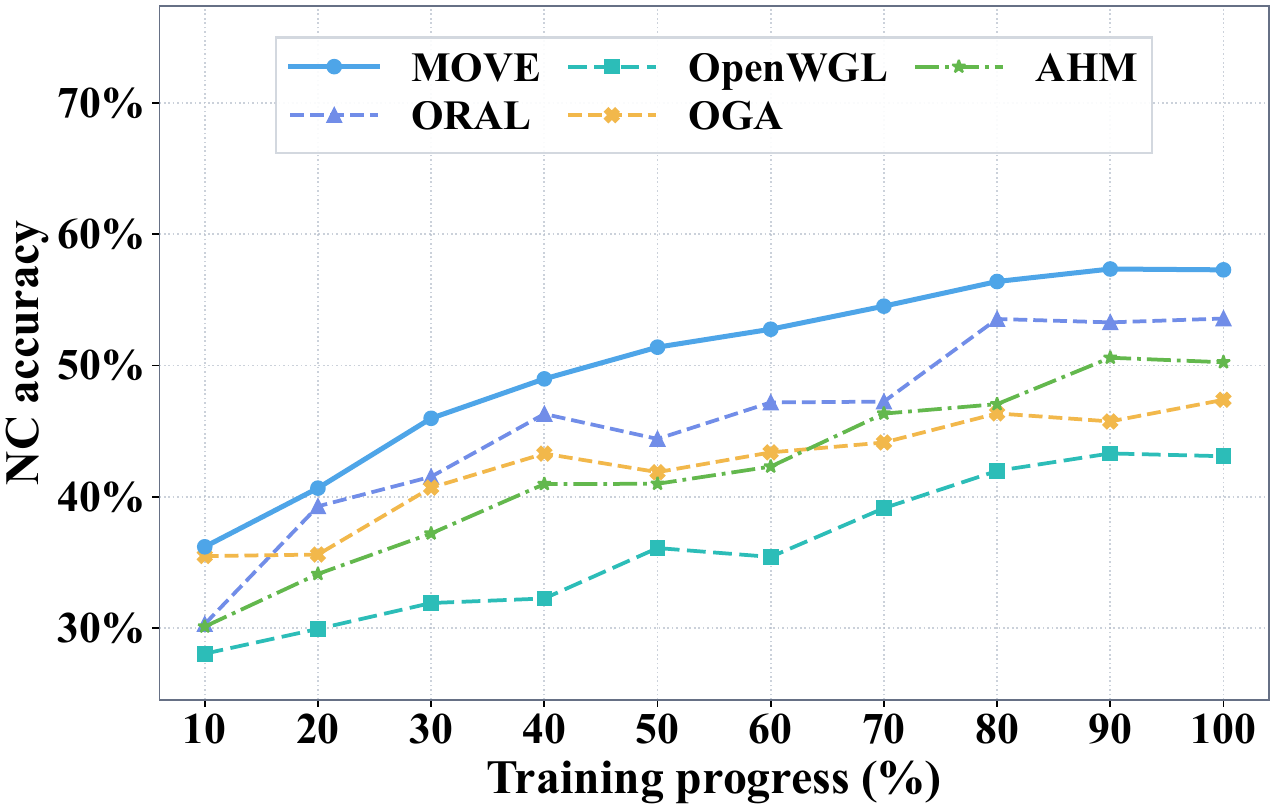}\\[-2pt]
    \small (b) Reddit-S.
\end{minipage}

\caption{
\textbf{Convergence comparison.}
Post-annotation NC accuracy of MOVE and annotation-capable baselines at different normalized upstream training progress.
}
\label{fig:method_convergence}

\end{minipage}
\hfill
\begin{minipage}[t]{0.49\textwidth}
\centering

\begin{minipage}[t]{0.485\linewidth}
    \centering
    \includegraphics[width=0.98\linewidth]
    {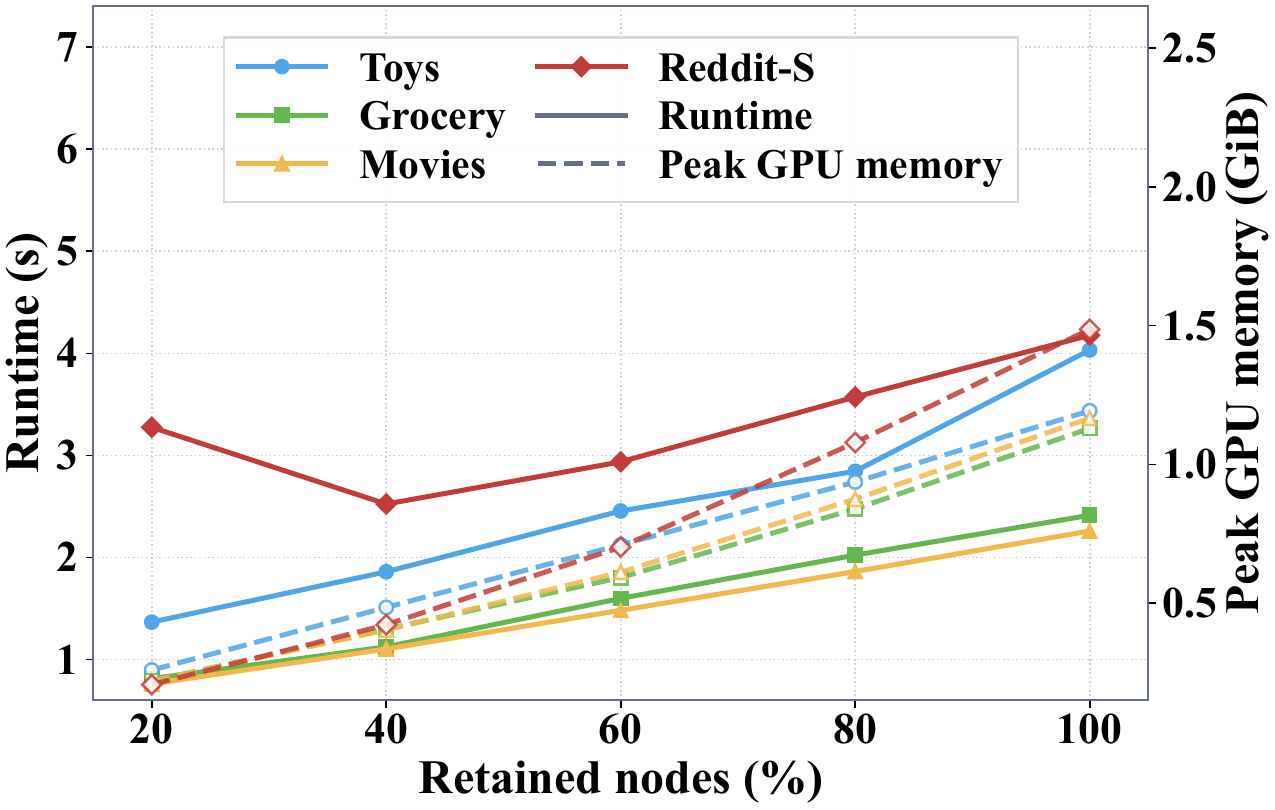}\\[-2pt]
    \small (a) Graph size.
\end{minipage}
\hfill
\begin{minipage}[t]{0.485\linewidth}
    \centering
    \includegraphics[width=0.98\linewidth]
    {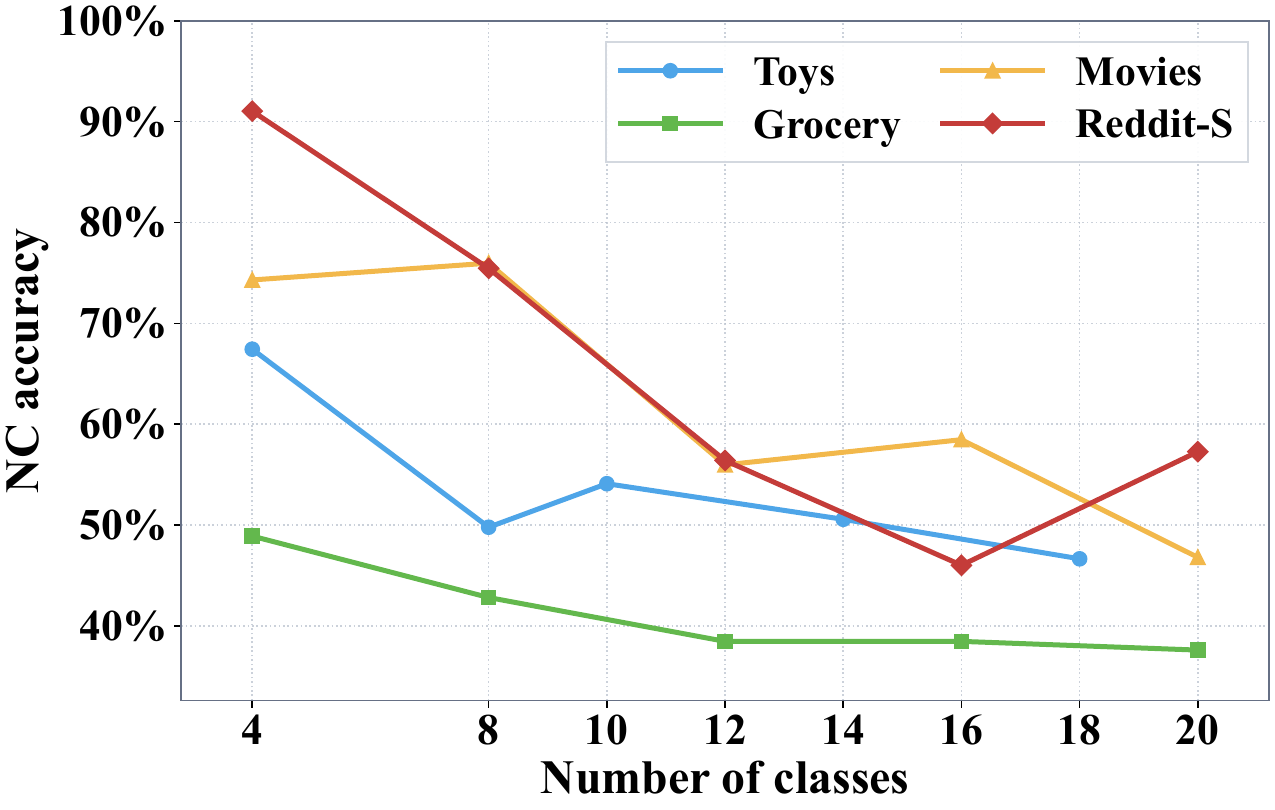}\\[-2pt]
    \small (b) Category size.
\end{minipage}

\caption{
\textbf{Scalability of MOVE.}
Runtime and peak GPU memory under increasing graph sizes, and post-annotation NC accuracy under increasing numbers of classes.
}
\label{fig:scalability}

\end{minipage}

\end{figure*}

To answer Q5, we examine training-progress convergence and evaluate scalability with respect to graph size and category size.

\textbf{Convergence Analysis.}
Figure~\ref{fig:method_convergence} shows that MOVE consistently achieves the highest post-annotation NC accuracy at every training-progress point. Its performance improves overall as training proceeds, whereas the baseline curves exhibit larger fluctuations or converge to lower Accuracy. These results demonstrate that MOVE learns useful unknown-class supervision throughout training and maintains a stable advantage over competing methods.

\textbf{Scalability Analysis.}
Figure~\ref{fig:scalability}(a) shows that runtime and peak GPU memory increase moderately with the retained-node ratio, remaining below 4.2 seconds and 1.5 GiB at full scale. Figure~\ref{fig:scalability}(b) shows that Accuracy generally decreases as the number of classes increases, reflecting the greater difficulty of a larger open-world label space. Nevertheless, MOVE retains 37.60\%--57.27\% Accuracy under the complete class configurations. Together, these results demonstrate that MOVE maintains manageable computational costs and effective downstream performance as the problem scale increases.

\section{Conclusion}

In this paper, we study open-world multimodal graph learning from the perspective of progressive knowledge evolution. Through empirical analysis, we reveal that unknown concepts require joint multimodal explanation, LLM-generated concepts need evidence-based verification, and naive expansion may introduce redundant categories. Based on these observations, we propose MOVE, a unified framework that identifies unresolved observations through uncertainty-aware multimodal coding, induces candidate concepts with multimodal LLMs and evidence grounding, and selectively expands the class space with reliable and non-redundant concepts. Extensive experiments demonstrate that MOVE effectively improves unknown-class recognition, semantic annotation, and downstream graph learning while maintaining a reliable and dynamic evolving class space.

\section*{AI Usage Disclosure}

Generative AI tools were used during the preparation of this manuscript for limited purposes, including language refinement, literature retrieval and discovery, and research ideation support. Specifically, AI tools assisted with improving writing clarity, organizing related work, exploring potential research directions, and discussing possible methodological designs and experimental considerations.

\bibliography{iclr2027_conference}

@inproceedings{stadler2021graph,
  title={Graph Neural Networks with Uncertainty Quantification for Out-of-Distribution Detection},
  author={Stadler, David and Charpentier, Bertrand and Z{\"u}gner, Daniel and G{\"u}nnemann, Stephan},
  booktitle={International Conference on Learning Representations (ICLR)},
  year={2021}
}

@inproceedings{wang2024openima,
  title={OpenIMA: Open-Set Graph Learning via Inter-Class Boundary Exploration},
  author={Wang, Yu and others},
  booktitle={Proceedings of the ACM Web Conference (WWW)},
  year={2024}
}

@inproceedings{fang2025graphgpto,
  title={GraphGPT-O: Multimodal Graph Learning with Large Language Models},
  author={Fang, Tao and others},
  booktitle={Proceedings of the ACM Multimedia Conference},
  year={2025}
}

@article{liu2026planet,
  title={PLANET: A Multimodal Graph Foundation Model for Universal Representation Learning},
  author={Liu, Xinyu and others},
  journal={arXiv preprint arXiv:2601.xxxxx},
  year={2026}
}

@article{kipf2016semi,
  title={Semi-supervised classification with graph convolutional networks},
  author={Kipf, Thomas N and Welling, Max},
  journal={arXiv preprint arXiv:1609.02907},
  year={2016}
}

@inproceedings{yan2025graph,
  title={When graph meets multimodal: benchmarking and meditating on multimodal attributed graph learning},
  author={Yan, Hao and Li, Chaozhuo and Yin, Jun and Yu, Zhigang and Han, Weihao and Li, Mingzheng and Zeng, Zhengxin and Sun, Hao and Wang, Senzhang},
  booktitle={Proceedings of the 31st ACM SIGKDD Conference on Knowledge Discovery and Data Mining V. 2},
  pages={5842--5853},
  year={2025}
}

@inproceedings{bao2024graph,
  title={Graph Out-of-Distribution Detection Goes Neighborhood Shaping.},
  author={Bao, Tianyi and Wu, Qitian and Jiang, Zetian and Chen, Yiting and Sun, Jiawei and Yan, Junchi},
  booktitle={ICML},
  pages={2923--2943},
  year={2024}
}

@article{wu2023energy,
  title={Energy-based out-of-distribution detection for graph neural networks},
  author={Wu, Qitian and Chen, Yiting and Yang, Chenxiao and Yan, Junchi},
  journal={arXiv preprint arXiv:2302.02914},
  year={2023}
}

@inproceedings{jin2024beyond,
  title={Beyond the known: Novel class discovery for open-world graph learning},
  author={Jin, Yucheng and Xiong, Yun and Fang, Juncheng and Wu, Xixi and He, Dongxiao and Jia, Xing and Zhao, Bingchen and Yu, Philip S},
  booktitle={International Conference on Database Systems for Advanced Applications},
  pages={117--133},
  year={2024},
  organization={Springer}
}

@inproceedings{wu2020openwgl,
  title={Openwgl: Open-world graph learning},
  author={Wu, Man and Pan, Shirui and Zhu, Xingquan},
  booktitle={2020 IEEE international conference on data mining (icdm)},
  pages={681--690},
  year={2020},
  organization={IEEE}
}

@article{wen2025llms,
  title={When LLMs meet open-world graph learning: a new perspective for unlabeled data uncertainty},
  author={Wen, Yanzhe and Li, Xunkai and Zhang, Qi and Lei, Zhu and Zeng, Guang and Li, Rong-Hua and Wang, Guoren},
  journal={arXiv preprint arXiv:2505.13989},
  year={2025}
}

@article{constantinou2026out,
  title={Out of distribution detection with attention head masking for multimodal document classification},
  author={Constantinou, Christos and Ioannides, Georgios and Chadha, Aman and Elkins, Aaron and Simpson, Edwin},
  journal={Scientific Reports},
  volume={16},
  number={1},
  pages={2449},
  year={2026},
  publisher={Nature Publishing Group UK London}
}

@inproceedings{ma2026coarse,
  title={Coarse-to-fine open-set graph node classification with large language models},
  author={Ma, Xueqi and Ma, Xingjun and Erfani, Sarah Monazam and Mandic, Danilo and Bailey, James},
  booktitle={Proceedings of the AAAI Conference on Artificial Intelligence},
  volume={40},
  number={43},
  pages={36714--36722},
  year={2026}
}

@inproceedings{xi2026topology,
  title={Topology-Aware Neural Collapse for Generalized Category Discovery on Graphs},
  author={Xi, Xuanzhi and Zhang, Zhong and Wang, Hongliang and Zhang, Tongze and hu, hongchao and Yang, Qinli and Shao, Junming},
  booktitle={Proceedings of the 32nd ACM SIGKDD Conference on Knowledge Discovery and Data Mining V. 2},
  pages={5579--5590},
  year={2026}
}
\bibliographystyle{iclr2027_conference}

\appendix

\section{Empirical Study in Detail}
\label{app:empirical_details}

This section specifies the diagnostic study in
Sec.~\ref{sec:empirical_study}. The study is designed to isolate the three
failure modes that motivate MOVE; it is not used to train MOVE or select its
hyperparameters. Unless stated otherwise, all thresholds are calibrated using
known-class data and fixed before candidate evaluation on independent
unknown-class evidence.

\subsection{Shared Representations and Diagnostic Protocol}
\label{app:empirical_shared}

\paragraph{Multimodal representations.}
We use the released CLIP ViT-L/14 visual and textual features and normalize
each feature to unit length, yielding $\mathbf{z}_i^v$ and $\mathbf{z}_i^t$.
To obtain a topology-conditioned representation without introducing an
additional learned model, we first form
$\widetilde{\mathbf{z}}_i=\operatorname{norm}(\mathbf{z}_i^v+\mathbf{z}_i^t)$
and aggregate one-hop incoming neighbors with a self-loop:
\begin{equation}
\mathbf{z}_i^g
=
\operatorname{norm}
\left(
\frac{
\widetilde{\mathbf{z}}_i+
\sum_{j:(j,i)\in\mathcal{E}}\widetilde{\mathbf{z}}_j
}{
1+|\{j:(j,i)\in\mathcal{E}\}|
}
\right).
\label{eq:empirical_graph_feature}
\end{equation}
The evidence sources are therefore
$\mathcal{R}=\{v,t,g\}$, corresponding to image, text, and graph context.
Cosine distance is denoted by
$d(\mathbf{a},\mathbf{b})=1-\mathbf{a}^{\top}\mathbf{b}$ for unit vectors.
The modality-specific prototype of each known class is estimated using its
training nodes:
\begin{equation}
\mathbf{c}_k^r
=
\operatorname{norm}
\left(
\frac{1}{|\mathcal{T}_k|}
\sum_{i\in\mathcal{T}_k}\mathbf{z}_i^r
\right),
\qquad r\in\mathcal{R}.
\label{eq:empirical_known_prototype}
\end{equation}

\paragraph{Proposal and validation evidence.}
For Obs.~2 and Obs.~3, held-out unknown nodes in the validation and test splits
form disjoint proposal and validation pools, denoted by $P$ and $V$,
respectively. Unknown-class identities are not used for region discovery,
candidate generation, candidate selection, threshold calibration, or support
decisions; they are accessed only for post-hoc analysis.

We fuse the three evidence sources as
\begin{equation}
\mathbf{f}_i
=
\operatorname{norm}
(\mathbf{z}_i^v+\mathbf{z}_i^t+\mathbf{z}_i^g)
\label{eq:empirical_region_fusion}
\end{equation}
and cluster the proposal pool into five unresolved regions. Central proposal
nodes from each region form the candidate-generation subset $P_g$, while
nodes in $V$ are assigned to the corresponding frozen regions to form the
independent validation subset $V_g$. This construction separates the evidence
used to generate a candidate from the evidence used to test its
generalization.

\subsection{Obs.~1: Cross-Modal Explanation Gap}
\label{app:empirical_obs1}

For node $i$, the best joint explanation by the known dictionary and the best
explanation available from any individual evidence source are, respectively,
\begin{equation}
C_i^{\mathrm{joint}}
=
\min_k
\frac{1}{|\mathcal{R}|}
\sum_{r\in\mathcal{R}}
d(\mathbf{z}_i^r,\mathbf{c}_k^r),
\qquad
C_i^{\mathrm{single}}
=
\min_{r\in\mathcal{R},k}
d(\mathbf{z}_i^r,\mathbf{c}_k^r).
\label{eq:empirical_joint_single_cost}
\end{equation}
The cross-modal explanation gap reported in
Fig.~\ref{fig:empirical_gap} is
\begin{equation}
G_i
=
C_i^{\mathrm{joint}}
-
C_i^{\mathrm{single}}.
\label{eq:empirical_gap_detailed}
\end{equation}
A node is operationally defined as \emph{single-source familiar but jointly
unexplained} when
\begin{equation}
C_i^{\mathrm{single}}
\leq
\tau_{\mathrm{fam}},
\qquad
G_i
>
\tau_{\mathrm{gap}},
\label{eq:empirical_gap_event}
\end{equation}
where $\tau_{\mathrm{fam}}$ and $\tau_{\mathrm{gap}}$ are calibrated using
known-class nodes. This avoids selecting an absolute distance scale using
unknown labels.

Across all four datasets, this event contains 11.7\% of unknown test nodes
and 9.9\% of known test nodes. The corresponding unknown/known rates are
13.7\%/10.0\% on Toys, 13.4\%/9.4\% on Grocery, 15.2\%/10.1\% on Movies,
and 9.9\%/6.2\% on Reddit-S. The higher unknown-node rate is consistently
observed across the four datasets, showing that an observation may appear
familiar under an individual modality while remaining incompatible with the
known class space when all evidence sources are considered jointly. The main
text visualizes Movies and Grocery for clarity.

\subsection{Obs.~2: Multimodal Evidence and Linguistic Plausibility}
\label{app:empirical_obs2}

\paragraph{Candidate generation.}
For each unresolved region, Qwen3-VL-Flash receives representative textual
and visual evidence together with the names of known classes and generates
five concise candidate descriptions. Neither $V_g$ nor the identities of
unknown classes are provided to the MLLM. Candidate descriptions are encoded
using the CLIP ViT-L/14 text encoder and unit-normalized, yielding the
semantic vector $\mathbf{q}_{g,m}$ for candidate $m$. Graph context is
withheld from candidate generation and is used only as an additional source
of evidence during verification.

\paragraph{Linguistic plausibility.}
Linguistic plausibility is assessed by the MLLM using proposal evidence only
and is recorded before multimodal verification. A candidate is considered
highly plausible when it provides a coherent description of the displayed
proposal evidence. This judgment measures linguistic plausibility rather than
correctness or support from independent evidence.

\paragraph{Operational definition of potential semantic hallucination.}
In this work, we use \emph{potential semantic hallucination} to denote a candidate class description that appears linguistically plausible given the proposal evidence but is not consistently supported by independent multimodal evidence. Specifically, a highly plausible candidate is regarded as potentially hallucinated if it fails at least one of the support conditions in Eq.~\ref{eq:empirical_consistent_support}. This is an evidence-based operational definition rather than a claim of factual incorrectness: a failed candidate may be semantically inappropriate, overly broad, specific to the proposal observations, or insufficiently supported by the available evidence. Unknown-class identities are used only for post-hoc evaluation and are not involved in this determination.

\paragraph{Multimodal evidence support.}
For split $S\in\{P,V\}$, we summarize each evidence source in region $g$ and
measure its alignment with candidate $m$:
\begin{equation}
\overline{\mathbf{z}}_{g,S}^{r}
=
\operatorname{norm}
\left(
\frac{1}{|S_g|}
\sum_{i\in S_g}\mathbf{z}_i^r
\right),
\qquad
a_{g,m,S}^{r}
=
(\overline{\mathbf{z}}_{g,S}^{r})^{\top}\mathbf{q}_{g,m}.
\label{eq:empirical_candidate_alignment}
\end{equation}
The average multimodal alignment is
\begin{equation}
a_{g,m,S}
=
\frac{1}{3}
\sum_{r\in\mathcal{R}}
a_{g,m,S}^{r}.
\end{equation}

The source-specific support threshold is calibrated exclusively using known
validation nodes. Let $\mathcal{A}_{r}^{+}$ contain the alignments between
each known-class name and the matching class evidence under source $r$, and
let $\mathcal{A}_{r}^{-}$ contain the corresponding mismatched alignments.
We define
\begin{equation}
\tau_r
=
\frac{1}{2}
\left[
Q_{0.25}(\mathcal{A}_{r}^{+})
+
Q_{0.75}(\mathcal{A}_{r}^{-})
\right].
\label{eq:empirical_alignment_threshold}
\end{equation}

A candidate is consistently supported on independent evidence if
\begin{equation}
\begin{aligned}
a_{g,m,V}^{r}&\geq\tau_r
&&\text{for every }r\in\mathcal{R},\\
\operatorname{rank}_{m'}(a_{g,m',V}^{r})&\leq 2
&&\text{for every }r\in\mathcal{R},\\
a_{g,m,V}&\geq a_{g,m,P}-\epsilon,
&&\epsilon=0.015.
\end{aligned}
\label{eq:empirical_consistent_support}
\end{equation}
These conditions require the candidate to receive sufficient support from
all three evidence sources, remain among the strongest alternatives under
each source, and avoid a substantial proposal-to-validation degradation.
The rules are fixed before candidate evaluation and do not use unknown-class
identities.

In the Movies and Grocery results, 6 of the 31 highly plausible candidates
satisfy all three conditions, yielding the reported support rate of 19.4\%.
The corresponding support rates within the highly plausible subset are
18.75\% on Movies and 20.00\% on Grocery. These results show that linguistic
plausibility does not necessarily imply consistent support from multimodal
evidence or generalization beyond the observations used for candidate
generation. Movies and Grocery are visualized in the main text, while the
same evaluation protocol is applied to all four datasets.

\subsection{Obs.~3: Proposal Fit and Independent Expansion Evidence}
\label{app:empirical_obs3}

Obs.~3 examines whether reducing reconstruction cost on proposal evidence is
sufficient to justify class-space expansion. For each unresolved region, we
compare four candidate constructions:
\begin{enumerate}
    \item \textbf{$P$-selected}: the MLLM candidate with the largest mean
    proposal alignment $a_{g,m,P}$;
    \item \textbf{Generic}: a deliberately broad dataset-level description;
    \item \textbf{Swapped}: a candidate transferred from another unresolved
    region of the same dataset; and
    \item \textbf{Centroid}: a candidate constructed directly from the
    modality-specific proposal centroids without natural-language semantics.
\end{enumerate}
These controls respectively represent proposal-based selection, overly broad
semantics, region-mismatched semantics, and direct local fitting.

For each language-based construction, the text vector is initially shared
across evidence sources,
$\mathbf{h}_{g,0}^{r}=\mathbf{q}_g$. We additionally evaluate a locally
adapted version:
\begin{equation}
\mathbf{h}_{g,\alpha}^{r}
=
\operatorname{norm}
\left[
(1-\alpha)\mathbf{q}_g
+
\alpha\overline{\mathbf{z}}_{g,P}^{r}
\right],
\qquad
\alpha\in\{0,0.5\}.
\label{eq:empirical_candidate_adaptation}
\end{equation}
The Centroid construction is already fully determined by proposal evidence
and is therefore unchanged under this comparison.

Let the known-dictionary reconstruction cost be
\begin{equation}
C_i(\mathcal{K})
=
\min_{k\in\mathcal{K}}
\sum_{r\in\mathcal{R}}
d(\mathbf{z}_i^r,\mathbf{c}_k^r).
\label{eq:empirical_expansion_base_cost}
\end{equation}
After tentatively adding candidate
$h=(\mathbf{h}^{v},\mathbf{h}^{t},\mathbf{h}^{g})$, the reconstruction cost
becomes
\begin{equation}
C_i(\mathcal{K}\cup\{h\})
=
\min
\left\{
C_i(\mathcal{K}),
\sum_{r\in\mathcal{R}}
d(\mathbf{z}_i^r,\mathbf{h}^{r})
\right\}.
\label{eq:empirical_expansion_extended_cost}
\end{equation}
For $S\in\{P,V\}$, the relative compression gain is
\begin{equation}
\Delta_g^S(h)
=
100\times
\frac{
\frac{1}{|S_g|}
\sum_{i\in S_g}C_i(\mathcal{K})
-
\frac{1}{|S_g|}
\sum_{i\in S_g}C_i(\mathcal{K}\cup\{h\})
}{
\frac{1}{|S_g|}
\sum_{i\in S_g}C_i(\mathcal{K})
}.
\label{eq:empirical_relative_gain}
\end{equation}

The same candidate is evaluated on both the proposal region $P_g$ and the
corresponding independent validation region $V_g$. In the Movies and Toys
results, the locally adapted Generic control obtains an average compression
gain of 1.9\% on proposal evidence but only 0.8\% on independent validation
evidence. The average proposal-to-validation gain reductions are 6.36
percentage points on Movies and 5.61 percentage points on Toys. Error bars
report standard errors across the five unresolved regions within each
dataset.

These results show that additional representational capacity may improve
proposal-set reconstruction even when the candidate does not provide
transferable explanatory value. Therefore, proposal-set improvement alone
cannot serve as a reliable class-admission rule, motivating the independent
validation and selective-expansion mechanism in MOVE. Movies and Toys are
visualized in the main text, while the same comparison is conducted on all
four datasets.
\section{Methodology in Details}
\label{app:method_details}

This section provides the implementation details of MOVE and clarifies the complete data flow shown in Fig.~\ref{fig:framework}. The framework contains a multimodal graph encoding front-end followed by three core stages: \textbf{Uncertainty-Aware Multimodal Class Coding}, \textbf{Evidence-Grounded Class Induction}, and \textbf{Selective Class Expansion}. The encoding front-end converts raw image, text, and graph topology into modality-specific node representations. The three MOVE stages then determine whether a node can be assigned to the current class space, infer candidate classes for unresolved nodes, and selectively incorporate only those candidates that remain reliable under independent evidence.

\subsection{Multimodal Graph Encoding}
\label{app:multimodal_encoding}

\paragraph{Input representation.}
Let $\mathcal{G}=(\mathcal{V},\mathcal{E},\mathbf{X}^{v},\mathbf{X}^{t})$ denote a multimodal attributed graph with adjacency structure $\mathbf{A}$. Each node $i$ is associated with visual input $\mathbf{x}_{i}^{v}$ and textual input $\mathbf{x}_{i}^{t}$. We first encode the two node-level modalities independently:
\begin{equation}
\mathbf{e}_{i}^{v}=f_{v}(\mathbf{x}_{i}^{v}),
\quad
\mathbf{e}_{i}^{t}=f_{t}(\mathbf{x}_{i}^{t}),
\label{eq:app_modality_encoding}
\end{equation}
where $f_v$ and $f_t$ denote the visual and textual encoders, respectively. Both representations are projected to a shared $d$-dimensional latent space before subsequent processing.

\paragraph{Topology-conditioned graph context.}
The graph branch is not treated as an independent raw modality. Instead, visual and textual features are first fused into an initial node representation,
\begin{equation}
\mathbf{h}_{i}^{(0)}
=
f_{\mathrm{fuse}}
\left(
[\mathbf{e}_{i}^{v}\Vert\mathbf{e}_{i}^{t}]
\right),
\label{eq:app_multimodal_fusion}
\end{equation}
which is propagated over the observed graph topology:
\begin{equation}
\mathbf{e}_{i}^{g}
=
\operatorname{GNN}
\left(
\mathbf{A},
\{\mathbf{h}_{j}^{(0)}\}_{j\in\mathcal{V}}
\right)_{i}.
\label{eq:app_graph_encoding}
\end{equation}
Here, $\mathbf{e}_{i}^{g}$ represents topology-conditioned multimodal context rather than topology alone. MOVE therefore retains three complementary representations for each node: $\mathbf{e}_{i}^{v}$ describing its visual content, $\mathbf{e}_{i}^{t}$ describing its textual attributes, and $\mathbf{e}_{i}^{g}$ describing its neighborhood-aware graph context. These representations form the input to uncertainty-aware class coding.

\subsection{Uncertainty-Aware Multimodal Class Coding}
\label{app:uncertainty_class_coding}

\paragraph{Modality-specific distributional encoding.}
A deterministic embedding does not indicate how reliable a modality is for a particular node. MOVE therefore maps the three representations to separate probability distributions. Each modality $r\in\{v,t,g\}$ is equipped with an independent trainable distribution head $\phi_r$:
\begin{equation}
\boldsymbol{\mu}_{i}^{r},
\mathbf{s}_{i}^{r}
=
\phi_r(\mathbf{e}_{i}^{r}),
\quad
\boldsymbol{\sigma}_{i}^{r}
=
\sigma_{\min}
+
\operatorname{softplus}(\mathbf{s}_{i}^{r}),
\quad
q_i^{r}(\mathbf{z})
=
\mathcal{N}
\left(
\boldsymbol{\mu}_{i}^{r},
\operatorname{diag}[(\boldsymbol{\sigma}_{i}^{r})^2]
\right).
\label{eq:app_distribution_encoding}
\end{equation}
The variance floor $\sigma_{\min}>0$ prevents numerical collapse. Thus, $\boldsymbol{\mu}_{i}^{r}$ represents the semantic location of modality $r$, whereas $\boldsymbol{\sigma}_{i}^{r}$ describes uncertainty along each latent dimension. The resulting $q_i^{v}$, $q_i^{t}$, and $q_i^{g}$ correspond to visual, textual, and graph-context evidence distributions, respectively.

\paragraph{Uncertainty-aware reliability.}
For a diagonal Gaussian, we compute its differential entropy as
\begin{equation}
\mathcal{H}(q_i^{r})
=
\frac{1}{2}
\sum_{d=1}^{D}
\log
\left(
2\pi e(\sigma_{i,d}^{r})^2
\right),
\quad
\omega_i^{r}
=
\frac{
\exp[-\mathcal{H}(q_i^{r})/\tau_{\omega}]
}{
\sum_{s\in\{v,t,g\}}
\exp[-\mathcal{H}(q_i^{s})/\tau_{\omega}]
}.
\label{eq:app_reliability}
\end{equation}
A modality with lower uncertainty therefore receives a larger reliability weight. The weights are node-specific and satisfy $\sum_r\omega_i^r=1$, allowing MOVE to reduce the effect of noisy or ambiguous modalities without manually selecting a dominant information source.

\paragraph{Existing-class representations.}
Each existing class $k$ is represented in the same latent space by a diagonal Gaussian
\begin{equation}
p_k(\mathbf{z})
=
\mathcal{N}
\left(
\boldsymbol{\mu}_{k},
\operatorname{diag}[\boldsymbol{\sigma}_{k}^{2}]
\right),
\quad
k=1,\ldots,K_t.
\label{eq:app_class_distribution}
\end{equation}
The class distributions are initialized from known-class training observations and optimized jointly with the trainable encoding and distributional heads under the coding objective in Eq.~\ref{eq:coding_training}. The class prior $\pi_k$ reflects the relative support of class $k$ in the known training data and satisfies $\sum_k\pi_k=1$.

\paragraph{Multimodal class matching.}
For every node--class pair $(i,k)$, MOVE measures two complementary quantities. The first evaluates whether each modality agrees with class $k$, while the second captures disagreement among the three evidence sources:
\begin{equation}
E_{ik}
=
\underbrace{
\sum_{r\in\{v,t,g\}}
\omega_i^{r}
D_{\mathrm{SKL}}
(q_i^{r}\Vert p_k)
}_{\text{class compatibility}}
+
\underbrace{
\gamma
\sum_{r<s}
\omega_i^{r}\omega_i^{s}
D_{\mathrm{B}}
(q_i^{r},q_i^{s})
}_{\text{cross-modal disagreement}}.
\label{eq:app_multimodal_energy}
\end{equation}
The class-compatibility term is class-specific, whereas the disagreement term is shared across classes for the same node. Consequently, cross-modal disagreement does not determine \emph{which} known class is preferred; instead, it raises the overall cost of assigning an internally inconsistent observation to the existing class space. This property directly supports the setting in which a node may appear familiar in one modality but still be unreliable when all modalities are considered jointly.

\paragraph{Class assignment and coding cost.}
The class-specific costs are converted into normalized assignment probabilities:
\begin{equation}
a_{ik}
=
\frac{
\pi_k\exp(-E_{ik}/\tau_c)
}{
\sum_{l=1}^{K_t}
\pi_l\exp(-E_{il}/\tau_c)
},
\quad
\mathcal{R}_i(\mathcal{C}_t)
=
-\tau_c
\log
\sum_{k=1}^{K_t}
\pi_k\exp(-E_{ik}/\tau_c).
\label{eq:app_assignment_coding}
\end{equation}
The assignment $a_{ik}$ answers which existing class best matches the node, whereas $\mathcal{R}_i(\mathcal{C}_t)$ measures whether the current class space as a whole can represent it. A small coding cost indicates that at least one existing class provides a compatible explanation; a large coding cost indicates that none of the current classes provides sufficient support.

\paragraph{Known-reference calibration.}
Raw coding costs are not directly comparable across datasets or classes. We therefore calibrate them using a held-out set of trusted known-class nodes. Let
\begin{equation}
\hat{c}_i=\arg\max_k a_{ik},
\quad
\mathcal{D}_{K}(\hat{c}_i)
=
\left\{
\mathcal{R}_j(\mathcal{C}_t)
\mid
j\in\mathcal{V}^{K}_{\mathrm{cal}},
y_j=\hat{c}_i
\right\}
\label{eq:app_calibration_reference}
\end{equation}
denote the class-conditional reference distribution for the predicted class of node $i$. MOVE computes the finite-sample calibration score
\begin{equation}
p_i
=
\frac{
1+
\sum_{s\in\mathcal{D}_{K}(\hat{c}_i)}
\mathbb{I}
[s\geq\mathcal{R}_i(\mathcal{C}_t)]
}{
1+
|\mathcal{D}_{K}(\hat{c}_i)|
},
\quad
b_i=-\log_2 p_i,
\label{eq:app_calibration_score}
\end{equation}
and performs a three-way state assignment:
\begin{equation}
\operatorname{state}(i)
=
\begin{cases}
\mathrm{known}, & p_i>\alpha_K,\\
\mathrm{pending}, & \alpha_U<p_i\leq\alpha_K,\\
\mathrm{unknown}, & p_i\leq\alpha_U,
\end{cases}
\quad
\alpha_U<\alpha_K.
\label{eq:app_three_way_state}
\end{equation}
Known nodes continue to use the existing class assignment. Unknown nodes seed new-class discovery. Pending nodes never initiate new classes; they are retained only as optional supporting context when they are sufficiently close to an unknown group, which avoids converting borderline observations into spurious discoveries.

\subsection{Evidence-Grounded Class Induction}
\label{app:evidence_grounded_induction}

\paragraph{Fused evidence distribution.}
For each unknown node, the three modality-specific evidence distributions are fused with a reliability-weighted product of experts:
\begin{equation}
q_i(\mathbf{z})
=
\frac{1}{Z_i}
\prod_{r\in\{v,t,g\}}
[q_i^{r}(\mathbf{z})]^{\omega_i^{r}}.
\label{eq:app_poe}
\end{equation}
Because all experts are Gaussian, the fused distribution also has a closed-form Gaussian representation:
\begin{equation}
\boldsymbol{\Lambda}_i
=
\sum_{r}
\omega_i^{r}
\boldsymbol{\Lambda}_{i}^{r},
\quad
\boldsymbol{\mu}_i
=
\boldsymbol{\Lambda}_{i}^{-1}
\sum_{r}
\omega_i^{r}
\boldsymbol{\Lambda}_{i}^{r}
\boldsymbol{\mu}_{i}^{r},
\quad
\boldsymbol{\Lambda}_{i}^{r}
=
(\boldsymbol{\Sigma}_{i}^{r})^{-1}.
\label{eq:app_poe_closed_form}
\end{equation}
This fused distribution summarizes the node-level multimodal evidence used for grouping and subsequent class induction.

\paragraph{Unknown-node grouping.}
MOVE does not assume that the number of unseen classes is known in advance. We therefore group unknown nodes according to the Bhattacharyya distance between their fused evidence distributions:
\begin{equation}
\{S_g\}_{g=1}^{G}
=
\operatorname{AggCluster}
\left(
\{q_i:
\operatorname{state}(i)=\mathrm{unknown}\},
D_{\mathrm{B}},
\tau_{\mathrm{grp}}
\right).
\label{eq:app_unknown_grouping}
\end{equation}
We use agglomerative grouping with a distance threshold $\tau_{\mathrm{grp}}$, so the number of discovered groups $G$ is determined by the observations rather than supplied as an oracle. Nodes within the same group therefore have similar joint visual, textual, and structural evidence. Groups that contain too few high-confidence unknown nodes are deferred rather than immediately triggering class generation.

\paragraph{Proposal--validation split.}
Each sufficiently supported group is randomly divided into two disjoint subsets:
\begin{equation}
S_g
=
S_g^{P}\cup S_g^{V},
\quad
S_g^{P}\cap S_g^{V}=\emptyset.
\label{eq:app_proposal_validation_split}
\end{equation}
The proposal subset $S_g^{P}$ is used to generate and ground candidate classes. The validation subset $S_g^{V}$ is never exposed to the MLLM or to candidate construction and is reserved exclusively for the selective-expansion stage. This separation is essential because evaluating a candidate on the same observations used to generate it would systematically favor overfitted or overly specific classes.

\paragraph{Multimodal evidence package.}
For every proposal group, MOVE forms an evidence package $\mathcal{E}_g$ from representative nodes in $S_g^P$. The package contains the corresponding visual inputs, textual attributes, and compact graph-context summaries derived from their local neighborhoods. The graph summary includes structural and neighbor information available in the underlying dataset, while avoiding unavailable ground-truth labels of unknown nodes. The evidence package therefore exposes the MLLM to the same three information sources used by MOVE rather than relying on natural-language node descriptions alone.

\paragraph{Candidate class proposal.}
The multimodal LLM receives $\mathcal{E}_g$ and returns a set of $M$ concise candidate class descriptions:
\begin{equation}
\mathcal{H}_g
=
\{h_{g,m}\}_{m=1}^{M}.
\label{eq:app_candidate_set}
\end{equation}
These descriptions are treated strictly as proposals. The MLLM does not decide which candidate is accepted, nor does linguistic plausibility alone constitute evidence for expanding the class space.

\paragraph{Candidate encoding.}
Each natural-language candidate description is mapped back to the same semantic space used by the node representations. Specifically, we encode the candidate with the textual semantic encoder and the corresponding distribution head:
\begin{equation}
\mathbf{e}_{g,m}^{h}
=
f_t(h_{g,m}),
\quad
(\boldsymbol{\mu}_{g,m}^{h},
\boldsymbol{\sigma}_{g,m}^{h})
=
\phi_t(\mathbf{e}_{g,m}^{h}),
\quad
q_{g,m}^{h}
=
\mathcal{N}
\left(
\boldsymbol{\mu}_{g,m}^{h},
\boldsymbol{\Sigma}_{g,m}^{h}
\right).
\label{eq:app_candidate_encoding}
\end{equation}
The resulting $q_{g,m}^{h}$ is the \emph{candidate prior}: it expresses where the MLLM-generated semantic description lies in the shared latent space before observing the proposal-group evidence. It is therefore not yet the final class representation.

\paragraph{Evidence grounding.}

MOVE combines the language-derived candidate prior with the multimodal
evidence of the proposal nodes. Let
$\boldsymbol{\Lambda}_{g,m}^{h}
=(\boldsymbol{\Sigma}_{g,m}^{h})^{-1}$.
We assign each proposal observation the normalized confidence weight
\begin{equation}
w_i
=
\frac{
\exp[-\mathcal{H}(q_i)/\tau_{\nu}]
}{
\sum_{j\in S_g^{P}}
\exp[-\mathcal{H}(q_j)/\tau_{\nu}]
},
\qquad
\sum_{i\in S_g^P}w_i=1.
\label{eq:app_evidence_confidence}
\end{equation}

To represent the population distribution of the candidate class, rather than
the posterior uncertainty of its sample mean, we construct the moment-matched
group distribution
\begin{equation}
\begin{aligned}
\bar{\boldsymbol{\mu}}_g
&=
\sum_{i\in S_g^P}
w_i\boldsymbol{\mu}_i,\\
\bar{\boldsymbol{\Sigma}}_g
&=
\sum_{i\in S_g^P}
w_i
\left[
\boldsymbol{\Sigma}_i
+
(\boldsymbol{\mu}_i-\bar{\boldsymbol{\mu}}_g)
(\boldsymbol{\mu}_i-\bar{\boldsymbol{\mu}}_g)^{\top}
\right]
+
\varepsilon_{\Sigma}\mathbf I,\\
\bar{\boldsymbol{\Lambda}}_g
&=
\bar{\boldsymbol{\Sigma}}_g^{-1}.
\end{aligned}
\label{eq:app_pooled_evidence}
\end{equation}
The first term preserves within-node uncertainty, while the outer-product
term captures between-node population scatter. The ridge
$\varepsilon_{\Sigma}\mathbf I$ ensures invertibility.

The candidate prior and the pooled evidence distribution are combined by
\begin{equation}
\begin{aligned}
\boldsymbol{\Lambda}_{g,m}^{*}
&=
\frac{
\beta\boldsymbol{\Lambda}_{g,m}^{h}
+
\bar{\boldsymbol{\Lambda}}_g
}{\beta+1},\\
\boldsymbol{\mu}_{g,m}^{*}
&=
\left(
\beta\boldsymbol{\Lambda}_{g,m}^{h}
+
\bar{\boldsymbol{\Lambda}}_g
\right)^{-1}
\left(
\beta\boldsymbol{\Lambda}_{g,m}^{h}
\boldsymbol{\mu}_{g,m}^{h}
+
\bar{\boldsymbol{\Lambda}}_g
\bar{\boldsymbol{\mu}}_g
\right).
\end{aligned}
\label{eq:app_grounding}
\end{equation}

The grounded candidate is
\begin{equation}
p_{g,m}^{*}
=
\mathcal{N}
\left(
\boldsymbol{\mu}_{g,m}^{*},
(\boldsymbol{\Lambda}_{g,m}^{*})^{-1}
\right).
\label{eq:app_grounded_candidate}
\end{equation}
Because the evidence weights sum to one and the pooled precision is normalized
by $\beta+1$, the candidate covariance does not shrink merely because the
proposal group contains more observations.
\paragraph{Proposal-set support.}
Before entering the expansion stage, each grounded candidate is evaluated on the proposal subset:
\begin{equation}
u_{g,m}
=
\frac{1}{|S_g^{P}|}
\sum_{i\in S_g^{P}}
\left[
\mathcal{R}_{i}(\mathcal{C}_t)
-
\mathcal{R}_{i}
(\mathcal{C}_t\cup\{p_{g,m}^{*}\})
\right],
\quad
\rho_{g,m}
=
\frac{
\exp(u_{g,m}/\tau_h)
}{
\sum_j\exp(u_{g,j}/\tau_h)
}.
\label{eq:app_proposal_support}
\end{equation}
Here, $u_{g,m}$ measures how much the candidate reduces assignment cost on the observations that proposed it, while $\rho_{g,m}$ provides a normalized support ranking among candidates from the same group. Importantly, $\rho_{g,m}$ is not an admission probability: final acceptance depends only on independent validation and the selective-expansion criterion described below.

\subsection{Selective Class Expansion}
\label{app:selective_expansion}

\paragraph{Provisional class prior.}
Evaluating $\mathcal{R}_i(\mathcal{C}_t\cup\{p_{g,m}^{*}\})$ requires a provisional prior for the candidate class. To avoid favoring candidates solely through an arbitrarily large mixture weight, we initialize the candidate prior from the size of its supporting proposal group:
\begin{equation}
\pi_{g}^{\mathrm{tmp}}
=
\frac{|S_g^{P}|}
{N_{K}+|S_g^{P}|},
\quad
\pi_{k}^{\mathrm{tmp}}
=
(1-\pi_{g}^{\mathrm{tmp}})\pi_k,
\label{eq:app_candidate_prior}
\end{equation}
where $N_K$ denotes the effective number of known reference observations. All candidates generated from the same group therefore use the same provisional prior, ensuring that their comparison depends on their representations rather than manually assigned prior mass.

\paragraph{Held-out generalization gain.}
The validation subset $S_g^{V}$ is independent of candidate generation. For each validation node, we measure the reduction in coding cost after tentatively introducing candidate $p_{g,m}^{*}$:
\begin{equation}
\Delta_{i,g,m}
=
\mathcal{R}_{i}(\mathcal{C}_t)
-
\mathcal{R}_{i}
(\mathcal{C}_t\cup\{p_{g,m}^{*}\}).
\label{eq:app_validation_gain}
\end{equation}
Let $\widehat{\sigma}_{g,m}$ denote the sample standard deviation of $\{\Delta_{i,g,m}\}_{i\in S_g^V}$. MOVE uses a one-sided lower confidence bound:
\begin{equation}
G_{g,m}^{\mathrm{LCB}}
=
\frac{1}{|S_g^{V}|}
\sum_{i\in S_g^{V}}
\Delta_{i,g,m}
-
z_{1-\delta_{\mathrm{conf}}}
\frac{
\widehat{\sigma}_{g,m}
}{
\sqrt{|S_g^{V}|}
}.
\label{eq:app_lcb_gain}
\end{equation}
A candidate therefore receives a large generalization score only when it consistently improves nodes that were not used to create or ground it. Here $\delta_{\mathrm{conf}}\in(0,1)$ is the one-sided tail probability and $z_{1-\delta_{\mathrm{conf}}}$ is the corresponding standard-normal quantile; this symbol is distinct from the semantic-complexity weight $\alpha$ defined below.

\paragraph{Redundancy and complexity control.}
A candidate that closely overlaps an existing class should not be added merely because it reduces local assignment cost. We therefore define a soft redundancy score
\begin{equation}
R_{g,m}^{\mathrm{red}}
=
\sum_{k=1}^{K_t}
\pi_k
\exp
\left(
-\frac{
D_{\mathrm{B}}(p_{g,m}^{*},p_k)
}{
\tau_{\Omega}
}
\right),
\label{eq:app_redundancy}
\end{equation}
which approaches one when the candidate is close to a high-probability existing class and decreases as it becomes semantically distinct. We further account for the effective complexity of introducing a new diagonal-Gaussian class:
\begin{equation}
C_{g,m}^{\mathrm{cmp}}
=
\frac{
d_{\mathrm{eff}}
}{
2|S_g^{P}|
}
\log(1+|S_g^{P}|),
\quad
\Omega_{g,m}
=
R_{g,m}^{\mathrm{red}}
+
\lambda_{\mathrm{cmp}}
C_{g,m}^{\mathrm{cmp}}.
\label{eq:app_redundancy_complexity}
\end{equation}
For a $d$-dimensional diagonal Gaussian, $d_{\mathrm{eff}}=2d$ when the mean and variance parameters are counted. The normalization by $|S_g^P|$ prevents a candidate supported by many observations from being penalized identically to one supported by only a few samples.

\paragraph{Operational hyperparameters in the robustness analysis.}
For reproducibility, we formally define the two implementation-level hyperparameters varied in Fig.~\ref{fig:robustness_hyperparameters}. Let $\mathbf z_{g,m}$ be the normalized embedding of candidate $h_{g,m}$ and let $\{\mathbf c_k\}_{k=1}^{K_t}$ be the normalized active-class prototypes. We compute the candidate residual and a dictionary-calibrated reference residual as
\begin{equation}
r_{g,m}
=
\min_k\|\mathbf z_{g,m}-\mathbf c_k\|_2^2,
\qquad
r_{\mathrm{ref}}
=
\operatorname*{median}_{k}
\min_{\ell\ne k}\|\mathbf c_k-\mathbf c_{\ell}\|_2^2,
\label{eq:app_semantic_residuals}
\end{equation}
and define the conditional semantic code length and its normalized multiplier by
\begin{equation}
\begin{aligned}
C_{g,m}^{\mathrm{sem}}
&=
\log_2\!\left(
1+\frac{r_{g,m}}{r_{\mathrm{ref}}+\varepsilon_{\mathrm{cmp}}}
\right),\\
\omega_{g,m}^{\mathrm{sem}}(\alpha)
&=
\operatorname{clip}
\left(
\frac{1+\alpha C_{g,m}^{\mathrm{sem}}}{1+\alpha},
\omega_{\min},
\omega_{\max}
\right),
\qquad \alpha\geq 0.
\end{aligned}
\label{eq:app_conditional_semantic_complexity}
\end{equation}
The implementation-level knowledge-complexity cost is $L_{g,m}^{\mathrm{cmp,op}}=\lambda \omega_{g,m}^{\mathrm{sem}}(\alpha)$, where $\lambda\geq0$ is the knowledge-complexity penalty shown on the horizontal axis of Fig.~\ref{fig:robustness_hyperparameters}(a--b), and $\alpha$ controls how strongly candidate-specific semantic distance modulates that penalty. This is the normalized operational instantiation of the complexity component in Eq.~\ref{eq:app_redundancy_complexity}; $\alpha=0$ recovers a constant unit complexity, and clipping prevents a single semantic residual from dominating the admission score.

An \emph{unknown anchor} is a high-confidence unknown node, i.e., a node with $p_i\leq\alpha_U$ in Eq.~\ref{eq:app_three_way_state}. For a proposal group $S_g$, define its unknown-anchor fraction and evidence gate as
\begin{equation}
f_g^{U}
=
\frac{1}{|S_g|}
\sum_{i\in S_g}
\mathbf{1}\{p_i\leq\alpha_U\},
\qquad
\operatorname{gate}_{\eta}(S_g)
=
\mathbf{1}\{f_g^{U}\geq\eta\},
\quad 0\leq\eta\leq1.
\label{eq:app_unknown_anchor_gate}
\end{equation}
Only groups passing this gate are eligible to trigger candidate admission; otherwise they are deferred until further evidence arrives. The robustness analysis varies $\eta=\texttt{min\_unknown\_anchor\_fraction}$ and uses $\eta=0.5$ by default. Thus, increasing $\eta$ requires a larger proportion of high-confidence unknown evidence in each proposal group.

\paragraph{Interference with existing classes.}
A valid new class should also avoid destabilizing decisions for already known observations. Let $\mathcal{B}_{\mathrm{old}}$ be a class-balanced buffer of trusted known nodes. Denote their class-assignment vectors before and after adding candidate $p_{g,m}^{*}$ by $\mathbf{a}_{i}^{-}\in\mathbb{R}^{K_t}$ and $\mathbf{a}_{i,g,m}^{+}\in\mathbb{R}^{K_t+1}$, respectively. We pad the original assignment by one zero entry and define
\begin{equation}
\Xi_{g,m}
=
\frac{1}{|\mathcal{B}_{\mathrm{old}}|}
\sum_{i\in\mathcal{B}_{\mathrm{old}}}
D_{\mathrm{JS}}
\left(
[\mathbf{a}_{i}^{-},0]
\Vert
\mathbf{a}_{i,g,m}^{+}
\right).
\label{eq:app_interference}
\end{equation}
A large $\Xi_{g,m}$ indicates that introducing the candidate substantially changes the predictions of known nodes and therefore risks damaging previously learned class boundaries.

\paragraph{Final admission decision.}
MOVE combines held-out improvement, redundancy/complexity, and old-class interference:
\begin{equation}
\mathcal{A}_{g,m}
=
G_{g,m}^{\mathrm{LCB}}
-
\lambda_{\Omega}\Omega_{g,m}
-
\lambda_{\Xi}\Xi_{g,m},
\quad
m^{*}
=
\arg\max_m
\mathcal{A}_{g,m}.
\label{eq:app_admission}
\end{equation}
The selected candidate is accepted only if its net admission score exceeds a conservative threshold $\tau_A$:
\begin{equation}
\mathcal{C}_{t+1}
=
\begin{cases}
\mathcal{C}_{t}
\cup
\{
(h_{g,m^{*}},
\pi_{g,m^{*}},
p_{g,m^{*}}^{*})
\},
&
\mathcal{A}_{g,m^{*}}
\geq
\tau_A,
\\
\mathcal{C}_{t},
&
\mathcal{A}_{g,m^{*}}
<
\tau_A.
\end{cases}
\label{eq:app_class_update}
\end{equation}
We use $\tau_A=0$ as the default decision rule, recovering the positive-net-gain criterion in the main text. The semantic description $h_{g,m^{*}}$ is retained as the generated class label used for unknown-class annotation, while $p_{g,m^{*}}^{*}$ becomes the distributional representation used for subsequent class assignment. The new class prior $\pi_{g,m^{*}}$ is initialized from its supporting group size and all class priors are renormalized after admission.

\paragraph{Sequential expansion.}
When several unknown groups are discovered at the same stage, candidates are processed in descending order of their admission scores. Once a class is accepted, $\mathcal{C}_t$ is immediately updated before evaluating subsequent groups. Consequently, later candidates are also compared against classes accepted earlier in the same expansion stage, reducing the possibility that two independently discovered groups introduce duplicate classes.

\subsection{Training and Deployment Procedure}
\label{app:training_deployment}

\paragraph{Known-class training.}
The trainable modality projections, graph encoder, distribution heads, and known-class representations are optimized only with labeled known-class training nodes. The coding loss in Eq.~\ref{eq:coding_training} encourages the correct known class to receive lower multimodal assignment cost than competing classes. No unknown-class labels are used for training the coding model.

\paragraph{Open-world deployment.}
At deployment, MOVE applies the trained coding model to unlabeled nodes and performs the three-way known/pending/unknown decision. Only high-confidence unknown nodes seed new-class discovery. Candidate class generation and grounding use the proposal subsets, while admission uses the disjoint validation subsets and the trusted old-class buffer. Unknown ground-truth labels are never accessed during proposal, grounding, or expansion.

\paragraph{Information flow across stages.}
The complete execution path is therefore
\[
(\mathbf{x}_i^{v},\mathbf{x}_i^{t},\mathbf{A})
\rightarrow
(\mathbf{e}_i^{v},\mathbf{e}_i^{t},\mathbf{e}_i^{g})
\rightarrow
(q_i^{v},q_i^{t},q_i^{g})
\rightarrow
\operatorname{state}(i)
\rightarrow
S_g
\rightarrow
(S_g^{P},S_g^{V})
\rightarrow
\mathcal{H}_g
\rightarrow
p_{g,m}^{*}
\rightarrow
\mathcal{A}_{g,m}
\rightarrow
\mathcal{C}_{t+1}.
\]
This data flow corresponds directly to Fig.~\ref{fig:framework}: multimodal graph encoding produces visual, textual, and topology-conditioned representations; uncertainty-aware class coding identifies unresolved nodes; evidence-grounded induction converts grouped unknown observations into grounded class candidates; and selective expansion determines which candidates are sufficiently general, non-redundant, and non-interfering to enter the evolving class space.

\section{Theoretical Analysis}
\label{sec:theory}

In this section, we analyze the theoretical property of MOVE's
counterfactual knowledge evolution. Specifically, we show that the proposed
compression-based admission criterion can distinguish reusable concepts from
spurious explanations that only improve local fitting.

\subsection{Consistency of Compression-Based Concept Admission}

Let $\mathcal{K}$ denote the current concept codebook and $p$ be a candidate
concept generated from an unresolved region. We define the population
compression gain of $p$ as

\begin{equation}
\Delta(p)
=
\mathbb{E}_{x\sim\mathcal{D}_{u}}
\left[
R(x;\mathcal{K})
-
R(x;\mathcal{K}\cup\{p\})
\right],
\end{equation}

where $\mathcal{D}_{u}$ denotes the distribution of unresolved observations
and $R(\cdot)$ is the conditional coding cost defined by MOVE.

The ideal admission rule is to accept $p$ when its expected explanatory gain
exceeds the additional knowledge complexity:

\begin{equation}
A^{*}(p)
=
\Delta(p)-\lambda_{\Omega}\Omega(p)
-\lambda_{\Xi}\Xi(p)>0 .
\end{equation}

However, MOVE only observes finite validation evidence. Given a validation set
$\mathcal{S}^{V}$, the empirical admission score is

\begin{equation}
\hat{A}(p)
=
\frac{1}{|\mathcal{S}^{V}|}
\sum_{x_i\in\mathcal{S}^{V}}
\left(
R(x_i;\mathcal{K})
-
R(x_i;\mathcal{K}\cup\{p\})
\right)
-\lambda_{\Omega}\Omega(p)
-\lambda_{\Xi}\Xi(p).
\end{equation}

\textbf{Theorem 1 (Counterfactual Admission Consistency).}
Assume that the coding gain variable
\[
Z_p(x)=R(x;\mathcal{K})-R(x;\mathcal{K}\cup\{p\})
\]
is bounded as $|Z_p(x)|\leq M$, and the validation samples are independently
drawn from $\mathcal{D}_{u}$. For a candidate concept $p$, if its population
admission margin satisfies

\begin{equation}
|A^{*}(p)|>\epsilon ,
\end{equation}

then the probability that MOVE makes an incorrect admission decision satisfies

\begin{equation}
\Pr
\left(
\operatorname{sign}(\hat{A}(p))
\neq
\operatorname{sign}(A^{*}(p))
\right)
\leq
2
\exp
\left(
-\frac{|\mathcal{S}^{V}|\epsilon^{2}}
{2M^{2}}
\right).
\end{equation}

Consequently, as the validation evidence increases,

\begin{equation}
\lim_{|\mathcal{S}^{V}|\rightarrow\infty}
\Pr
\left(
\operatorname{sign}(\hat{A}(p))
=
\operatorname{sign}(A^{*}(p))
\right)
=1 .
\end{equation}

\textbf{Proof.}
Since the validation samples are independently drawn, the empirical
compression gain is an unbiased estimator of the population compression gain:

\begin{equation}
\mathbb{E}[\hat{A}(p)]
=
A^{*}(p).
\end{equation}

The complexity and interference terms are deterministic with respect to the
candidate concept. Therefore, an incorrect admission occurs only when

\begin{equation}
|\hat{A}(p)-A^{*}(p)|>\epsilon .
\end{equation}

Applying Hoeffding's inequality to the bounded random variable $Z_p(x)\in[-M,M]$, whose interval length is $2M$, gives

\begin{equation}
\Pr\left(
|\hat{A}(p)-A^{*}(p)|>\epsilon
\right)
\leq
2
\exp\left(
-\frac{|\mathcal{S}^{V}|\epsilon^{2}}{2M^{2}}
\right).
\end{equation}

Therefore, the probability of accepting a spurious concept or rejecting a
genuine concept decreases exponentially with the amount of independent
validation evidence. This proves the consistency of the proposed
counterfactual admission criterion.
\qed

\subsection{Consistency of Uncertainty-Aware Multimodal Coding}

\label{sec:theory_coding}

MOVE identifies unknown observations through Uncertainty-Aware multimodal coding.
We analyze when the proposed coding objective can distinguish observations that
are explainable by existing concepts from those requiring new knowledge.

Let $p_k$ denote an existing concept and
$q_i^r$ denote the evidence distribution of node $i$ under modality
$r\in\{v,t,g\}$. The concept-conditioned energy is defined as

\begin{equation}
E_{ik}
=
\sum_{r}\omega_i^r
D(q_i^r\|p_k)
+
\gamma
\sum_{r<s}
\omega_i^r\omega_i^s
D_B(q_i^r,q_i^s).
\end{equation}

The corresponding coding cost is

\begin{equation}
R_i(\mathcal K)
=
-\tau_c
\log
\sum_k
\pi_k
\exp(-E_{ik}/\tau_c).
\end{equation}

We consider two types of observations. An explainable observation satisfies
that there exists an existing concept $p_k$ such that all evidence sources
remain close to this concept. In contrast, an unexplained observation has a
non-zero multimodal mismatch margin:

\begin{equation}
\min_k E_{ik}
\geq
\delta ,
\end{equation}

where $\delta$ measures the irreducible explanation gap between the observation
and the current knowledge.

\textbf{Theorem 2 (Multimodal Explanation Separation).}
Let $E_{ik}^{(m)}$ be the energy obtained from a fixed set of $m$ modality terms, using the same nonnegative reliability coefficients as the full model. Suppose that adding modality $r'$ contributes the nonnegative increment
\begin{equation}
B_{ik}^{r'}
=
E_{ik}^{(m+1)}-E_{ik}^{(m)}
\geq 0
\label{eq:theory_incremental_energy}
\end{equation}
to class $k$. Define the assignment distribution based on the first $m$ terms,
\begin{equation}
a_{ik}^{(m)}
=
\frac{
\pi_k\exp(-E_{ik}^{(m)}/\tau_c)
}{
\sum_{\ell}\pi_{\ell}
\exp(-E_{i\ell}^{(m)}/\tau_c)
}.
\label{eq:theory_partial_assignment}
\end{equation}
If an unexplained observation satisfies $B_{ik}^{r'}\geq\delta_{r'}>0$ for a set of classes carrying total mass at least $\rho$ under $a_i^{(m)}$, then
\begin{equation}
R_i^{(m+1)}(\mathcal K)-R_i^{(m)}(\mathcal K)
\geq
-\tau_c\log\!\left[
1-\rho\left(1-e^{-\delta_{r'}/\tau_c}\right)
\right].
\label{eq:theory_unexplained_lower_bound}
\end{equation}
In particular, if every existing class has incremental mismatch at least $\delta_{r'}$, then the increase is at least $\delta_{r'}$. Conversely, if an explainable observation satisfies $0\leq B_{ik}^{r'}\leq\epsilon_{r'}$ for every class with nonzero assignment mass, then
\begin{equation}
0
\leq
R_i^{(m+1)}(\mathcal K)-R_i^{(m)}(\mathcal K)
\leq
\epsilon_{r'}.
\label{eq:theory_explainable_upper_bound}
\end{equation}
Hence the added modality increases the unexplained--explainable coding-cost gap by at least the right-hand side of Eq.~\ref{eq:theory_unexplained_lower_bound} minus $\epsilon_{r'}$ whenever this quantity is positive. The same bounds hold in expectation when the stated increment conditions hold almost surely.

\textbf{Proof.}
Using Eq.~\ref{eq:theory_partial_assignment} and $E_{ik}^{(m+1)}=E_{ik}^{(m)}+B_{ik}^{r'}$, the change in coding cost has the exact form
\begin{align}
R_i^{(m+1)}-R_i^{(m)}
&=
-\tau_c\log
\frac{
\sum_k\pi_k
\exp[-(E_{ik}^{(m)}+B_{ik}^{r'})/\tau_c]
}{
\sum_k\pi_k\exp(-E_{ik}^{(m)}/\tau_c)
}\\
&=
-\tau_c\log
\sum_k a_{ik}^{(m)}
\exp(-B_{ik}^{r'}/\tau_c).
\label{eq:theory_exact_increment}
\end{align}
Because $B_{ik}^{r'}\geq0$, the weighted exponential sum in Eq.~\ref{eq:theory_exact_increment} is at most one, proving the nonnegative lower bound. Let $\mathcal I$ be the set of classes for which $B_{ik}^{r'}\geq\delta_{r'}$ and let $\sum_{k\in\mathcal I}a_{ik}^{(m)}\geq\rho$. Then
\begin{equation}
\sum_k a_{ik}^{(m)}e^{-B_{ik}^{r'}/\tau_c}
\leq
(1-\rho)+\rho e^{-\delta_{r'}/\tau_c}.
\end{equation}
Since $-\log(\cdot)$ is decreasing, substituting this upper bound into Eq.~\ref{eq:theory_exact_increment} proves Eq.~\ref{eq:theory_unexplained_lower_bound}. Setting $\rho=1$ reduces its right-hand side to $\delta_{r'}$.

For an explainable observation, $B_{ik}^{r'}\leq\epsilon_{r'}$ implies $e^{-B_{ik}^{r'}/\tau_c}\geq e^{-\epsilon_{r'}/\tau_c}$ for every class with nonzero assignment mass. Therefore,
\begin{equation}
\sum_k a_{ik}^{(m)}e^{-B_{ik}^{r'}/\tau_c}
\geq e^{-\epsilon_{r'}/\tau_c},
\end{equation}
and Eq.~\ref{eq:theory_exact_increment} gives the upper bound $R_i^{(m+1)}-R_i^{(m)}\leq\epsilon_{r'}$. Subtracting this upper bound from the unexplained-observation lower bound yields the claimed separation result. Finally, almost-sure inequalities are preserved under expectation, completing the proof.
\qed

\section{Experiment in Details}
\label{sec:experiment_details}

\textbf{Datasets.}
We evaluate MOVE on four multimodal attributed graph datasets, including Toys, Grocery, Movies, and Reddit-S~\cite{yan2025graph}. These datasets represent different real-world scenarios with heterogeneous node attributes and relational structures. Each node is associated with multimodal information, including textual attributes, visual content, and graph connections. Toys and Grocery represent product-centered graphs where nodes contain item descriptions, images, and user interaction structures. Movies represents a multimedia recommendation graph containing textual metadata, visual information, and relational links among entities. Reddit-S represents a social graph where nodes are associated with semantic attributes and network structures. These datasets provide diverse evaluation environments for testing open-world class recognition and expansion.

\textbf{Open-world Protocol.}
Following the open-world graph learning setting, we divide the original class space into known and unknown categories. Specifically, 50\% of the classes are randomly selected as known classes, whose nodes are available during training, while the remaining 50\% are treated as unknown classes that only appear during evaluation. The model is trained exclusively with known-class nodes and is required to recognize known nodes, identify unknown nodes, infer their potential classes, and determine whether new classes should be incorporated into the evolving class space. For downstream evaluation, we compare three settings: the original graph without unknown-class annotation, the graph annotated by MOVE, and the graph annotated using ground-truth unknown labels.

\textbf{Baselines.}
We compare MOVE with representative methods from three categories. The first category contains OOD detection methods, including TopoOOD~\cite{bao2024graph} and GNN-Safe~\cite{wu2023energy}, which focus on identifying nodes outside the training distribution but do not explicitly generate new class descriptions. The second category contains structural open-world graph learning methods, including ORAL~\cite{jin2024beyond} and OpenWGL~\cite{wu2020openwgl}, which identify or organize unknown nodes. For semantic label quality evaluation, their outputs are converted into natural-language labels using the common adaptation described below. The third category contains multimodal open-world methods, including OGA~\cite{wen2025llms} and AHM~\cite{constantinou2026out}, which leverage multimodal or LLM-based information for open-world recognition. We additionally compare with recent graph-based open-world methods, including CFC~\cite{ma2026coarse} and TopoNC~\cite{xi2026topology}, to evaluate the effectiveness of topology-aware unknown recognition. Methods without native class annotation or expansion capabilities are evaluated only on the tasks supported by their original designs.

\textbf{Semantic-label adaptation.}
Since OpenWGL and ORAL do not natively generate natural-language class labels, we adapt them to the UCA evaluation as follows. We retain the unknown-node predictions produced by each method, group the predicted unknown nodes according to graph connectivity, and summarize the representative multimodal evidence of each group into a concise semantic label using LLM. This adaptation is used only to provide the natural-language outputs required by the semantic label quality metrics and does not modify the original unknown-node recognition results.

\textbf{Evaluation Tasks.}
We evaluate MOVE from three perspectives following the complete open-world pipeline.

\emph{Unknown-Class Rejection (UCR).}
The first task evaluates whether the model can distinguish unknown nodes from known classes. We report known-class recognition accuracy and unknown-node rejection performance. Accuracy measures whether known categories are preserved after open-world learning, while Coverage and Precision evaluate the ability to identify unknown nodes.

\emph{Unknown-Class Annotation (UCA).}
The second task evaluates whether the discovered classes accurately describe previously unseen concepts. We measure the semantic similarity between generated class descriptions and ground-truth unknown labels using G-to-U similarity, where higher values indicate better alignment with true unknown categories. We also report K-to-G similarity, which measures the similarity between generated classes and known classes, where lower values indicate fewer redundant expansions.

\emph{Downstream Graph Learning.}
Finally, we evaluate whether the discovered classes provide useful information for downstream graph learning. We conduct node classification using a GCN~\cite{kipf2016semi} backbone and report classification accuracy. We also evaluate zero-shot modality matching to measure whether the generated classes preserve multimodal semantic information. For node classification, the graph without unknown-class annotation and the graph with ground-truth unknown labels are used as lower and upper bounds, respectively.

\textbf{Ablation Variants.}
We evaluate four functional components of MOVE. The corresponding ablation variants are defined as follows.

\emph{Cross-Modal Disagreement Modeling (CDM).}
CDM captures inconsistency among modality-specific evidence through the pairwise disagreement term in Eq.~\ref{eq:concept_energy}. The \emph{w/o CDM} variant removes this term, so that class coding depends only on the compatibility between individual modalities and existing classes.

\emph{Held-out Candidate Validation (HCV).}
HCV evaluates grounded candidate classes on the independent validation subset $S_g^V$. The \emph{w/o HCV} variant selects candidates using only the proposal subset $S_g^P$, without testing their generalization to held-out observations.

\emph{Graph-Context Encoding (GCE).}
GCE incorporates topology-conditioned representations into multimodal class coding and candidate induction. The \emph{w/o GCE} variant removes graph-context evidence and retains only visual and textual information.

\emph{Class-Space Update (CSU).}
CSU incorporates admitted candidate classes into the evolving class space $\mathcal{C}_{t+1}$. In the \emph{w/o CSU} variant, generated classes may be used to annotate the corresponding unknown nodes but are not incorporated into the class space for subsequent learning.

\end{document}